\documentclass[10pt,a4paper,logo]{paper}

\usepackage[all]{hypcap}

\usepackage[authoryear, round]{natbib}

\usepackage{nicefrac}
\usepackage{multirow}
\usepackage[capitalize,noabbrev]{cleveref}
\crefname{appendix}{Appendix}{Appendices}
\Crefname{appendix}{Appendix}{Appendices}
\usepackage{dsfont}
\usepackage{makecell}
\usepackage{subcaption}
\usepackage{bbm}
\usepackage{wrapfig}
\usepackage{mathtools,xparse}
\usepackage{array}
\usepackage{arydshln}
\usepackage{scalerel}
\usepackage{tablefootnote}
\usepackage{tocloft}
\usepackage{adjustbox}
\usepackage{setspace}
\usepackage[most,skins,theorems]{tcolorbox}
\usepackage{xspace}

\newboolean{showsection}
\setboolean{showsection}{true}
\makeatletter
\@namedef{ver@everyshi.sty}{}
\makeatother

\definecolor{custom_green}{rgb}{0.0, 0.5, 0.0}
\definecolor{custom_red}{rgb}{1.0, 0.01, 0.24}
\definecolor{custom_blue}{HTML}{C9DAF7}
\definecolor{custom_purple}{HTML}{D9D1E9}
\definecolor{title_blue}{HTML}{204899}
\definecolor{cite_blue}{HTML}{044dc1}
\definecolor{cite_purple}{HTML}{7406a7}

\hypersetup{
    colorlinks = true,
    citecolor = {cite_blue},
    linkcolor = {cite_purple},
    urlcolor = {cite_purple},
}

\usepackage{amsmath,amsfonts,bm}

\def\eqref#1{equation~\ref{#1}}

\def\1{\bm{1}}

\DeclareMathAlphabet{\mathsfit}{\encodingdefault}{\sfdefault}{m}{sl}
\SetMathAlphabet{\mathsfit}{bold}{\encodingdefault}{\sfdefault}{bx}{n}

\makeatletter
\def\mathcolor#1#{\@mathcolor{#1}}
\def\@mathcolor#1#2#3{%
  \protect\leavevmode
  \begingroup
    \color#1{#2}#3%
  \endgroup
}
\makeatother

\Crefformat{equation}{#2Eq.\;(#1)#3}
\Crefformat{figure}{#2Figure #1#3}
\Crefformat{assumption}{#2Assumption #1#3}
\Crefname{assumption}{Assumption}{Assumptions}

\usepackage{crossreftools}
\makeatletter
\renewcommand\footnoterule{%
  \kern 15\p@
  \hrule \@width 2in \kern 2.6\p@
  \vspace{4pt}
}
\makeatother

\newcommand{\modelname}{RunningTab\xspace}

\definecolor{mygray}{HTML}{F2F2F2}
\definecolor{citeblue}{HTML}{1668b0}

\newcommand{\std}[1]{\,{\scriptsize($\pm$#1)}}
\newcommand{\negdelta}[1]{\textcolor{red!70!black}{\,\scriptsize$-$#1}}

\title{RunningTab: Direct Workspace Interaction \\ with Environment-Side Tabs}

\reportnumber{}

\author[1]{Jinheon Baek}
\author[1]{Soyeong Jeong}
\author[1]{Yumin Choi}
\author[1]{Dongsu Han}
\author[1,2]{Sung Ju Hwang}

\affil[1]{KAIST}
\affil[2]{DeepAuto.ai}

\correspondingauthor{Correspondence to: \href{mailto:jinheon.baek@kaist.ac.kr,sungju.hwang@kaist.ac.kr}{\texttt{\{jinheon.baek, sungju.hwang\}@kaist.ac.kr}}.}


\begin{abstract}
Much knowledge work produces new deliverables from files a workspace already holds, and LLM agents are beginning to take such work over. Through direct corpus interaction, an agent can search and read any of those files from a terminal with no indexing, and producing a deliverable from many of them in this way is what we call \emph{direct workspace interaction} (DWI). Reaching the files, however, is only half the task: nothing keeps track of what the task asks for, what has been read, and what was listed but never opened, all of which slip through the context window without leaving a trace, so an agent may extract a figure and still deliver a report without it. To address this, we present \modelname, a framework that equips direct workspace interaction with an \emph{environment-side tab}: a per-task record of what the task still owes, kept by the environment alongside the agent. Specifically, the agent adds its requirements, while the environment records every file read as an excerpt with its provenance and every listed but unopened file as a candidate; the agent can then see each requirement beside its best-matching excerpts and top unopened candidates, resolve it against matching content or set it aside with a reason, and, should it try to finish with requirements still open, receive them in a finish check. We validate \modelname on three benchmarks with three LLMs, where it consistently outperforms plain DWI and baselines that keep the record in the model, while its tab usually holds the values a deliverable needs once seen.
\end{abstract}

\begin{document}

\maketitle

\section{Introduction}

Much of the work in an organization consists of producing new deliverables from files it already holds (the reports, spreadsheets, slide decks, and notes accumulated over years in a shared workspace)~\citep{TheAgentCompany, GDPval, WorkspaceBench}: a hiring summary is compiled from resumes and interview notes; a budget review is built from a year of monthly spreadsheets; a project proposal is drafted from meeting notes and a previous contract. Large Language Model (LLM) agents that operate a computer autonomously~\citep{ReAct, Toolformer, SWEAgent, TerminalBench} make it possible to hand such work over end to end, so that the agent itself locates the relevant files, reads them, and writes the deliverable. A workspace task, then, is not a single retrieval but an accumulation over a long horizon: the deliverable brings together content scattered across many files, and the agent must reach each of them before it can write.

To reach the existing files, \emph{direct corpus interaction}~\citep{DCI} opens the road: instead of a retriever over a prebuilt index~\citep{DPR, RAG}, the agent works in a terminal and applies standard command-line utilities (listing a directory, searching for a term across files, opening a document) directly to the raw files, so that any file in the workspace can be reached as it is, with no indexing or preprocessing. Subsequent work scales it to larger corpora~\citep{DRDCI}, synthesizes tasks for it~\citep{GrepSeek}, and studies how agents search under it~\citep{IsGrepAllYouNeed, KeywordSearch}. However, when the corpus is such a workspace and the goal is not a short answer to a question but a new deliverable assembled from many of its files, the agent has to work inside the workspace over many steps and hand in the deliverable at the end; we refer to this setting as \emph{direct workspace interaction} (DWI). Yet reaching the files is only half the task.

\begin{figure}[t!]
    \centering
    \includegraphics[width=\linewidth]{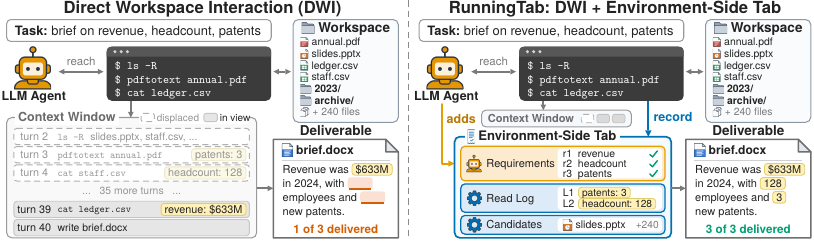}
    \vspace{-0.20in}
    \caption{Direct workspace interaction (DWI) and \modelname on an illustrative task: a brief that needs three facts scattered across a workspace. (Left) In DWI, the agent reaches every file through its terminal, but what it reads lives only in its context window: facts read early are displaced by later observations, and the brief is delivered without them. (Right) \modelname keeps an environment-side tab beside the agent: the agent adds the requirements of the task, the environment records what is read and what is listed but not opened, and the facts remain available until the deliverable is written.}
    \label{fig:concept}
    \vspace{-0.10in}
\end{figure}

The agent has full access to the workspace, but nothing keeps track of the task it is carrying out. What the task asks for, which files have been read, and which were listed but never opened pass through the context window and are soon displaced by new observations; nothing records which of them have been used, and nothing connects a requirement of the task to the file that satisfies it. As a consequence, an agent may open the right annual report, extract the figure it needs into its observations, and still deliver a report without that figure; or it may list a folder that contains the relevant file and never open it. What is missing, we argue, is a record of what the task still owes, kept by the environment and available to the agent whenever it needs it.

In this work, we present \textbf{\modelname}, a framework that equips direct workspace interaction with an \emph{environment-side tab}: a per-task running record that the environment maintains for the agent (\cref{fig:concept}). Specifically, the agent adds one \emph{requirement} per fact, figure, or element the task asks for, while the environment records every file read as an excerpt with its provenance and every file listed but not opened as a \emph{candidate}, further ranked by its relevance to the task. Because the tab keeps what would otherwise have passed out of view, the agent can consult it freely, returning to a stored excerpt or a listed file whenever it needs them instead of searching the workspace again. Furthermore, before resolving any requirement, the agent lists them and sees each open one beside its best-matching excerpts and its top unopened candidates, and a requirement is resolved only against an excerpt whose content matches it, or with a reason it cannot be met; should the agent attempt to finish with items open, the environment returns them once, in a \emph{finish check}. The tab thus runs alongside the task, from the first requirement the agent writes to the finish check, holding what the model would otherwise have to retain and pointing back to the files it came from.

We validate \modelname on three benchmarks~\citep{WorkspaceBench, TheAgentCompany, OfficeQAPro} with three LLMs, where it improves over plain DWI and baselines in which the model itself keeps track of the task, in every case. Ablations attribute most of this gain to what the environment captures on its own, the tab usually holds the values a deliverable needs once the agent has seen them, and the gain grows with how much the agent reads. These results affirm that direct workspace interaction needs more than access to the files: the terminal reaches them, and the environment-side tab keeps what the task owes until it is delivered.

\section{Related Work}
\label{sec:related_work}

\paragraph{Workspace Agents}
LLM agents that operate a computer have shown strong performance as software developers and terminal users~\citep{SWEAgent, OpenHands} and are evaluated in terminal and desktop environments~\citep{TerminalBench, OSWorld}. Building on these agents, recent benchmarks bring them to knowledge work: they pose tasks over realistic office and enterprise workspaces and judge the deliverable the agent produces~\citep{TheAgentCompany, GDPval, WorkspaceBench, OfficeBench, OfficeQAPro, TUABench}. However, these studies find that the deliverables often lack content the workspace holds: \citet{TheAgentCompany} attribute failures in part to a lack of ability to understand documents, and \citet{WorkspaceBench} name heterogeneous file understanding and lineage tracing as the bottlenecks. These diagnoses concern reaching and reading the files, whereas our own analysis (discussed in \cref{sec:dwi} with \cref{fig:diagnosis}) takes a different lens and finds a loss that these diagnoses do not cover: content from files the agent had already opened, which appeared in its observations yet is missing from the deliverable. In this work, we address this part directly, by keeping beside the agent a record of what the task still owes and of what has been read, so that content once reached is not lost before delivery.

\paragraph{Direct Corpus Interaction}
Retrieval over a prebuilt index has long been the way a model reaches a corpus beyond its context~\citep{DPR, RAG}, first as a single retrieval step and then, in agentic search, as repeated queries to the same retriever over many rounds~\citep{SearchR1, SearchO1}. Direct corpus interaction departs from this design: rather than a retriever that decides what the agent sees, the agent is given the corpus itself and searches it with terminal tools~\citep{DCI}. In particular, Dr-DCI~\citep{DRDCI} lets the agent pull retrieved documents into a local workspace (where a workspace denotes the materialized set of pulled documents, whereas here it denotes the files an organization holds), GrepSeek~\citep{GrepSeek} trains compact agents for it, and related work compares terminal or keyword search with vector retrieval~\citep{IsGrepAllYouNeed, KeywordSearch} or builds on it~\citep{InteractionSpaces, RelevanceGuidingCorpusInteraction}. However, these works typically target a short answer to a question and improve how an agent reaches the corpus; to our knowledge, none of them records which file meets which requirement of the task. \modelname instead turns to what becomes of the content once it is reached: it changes nothing about how the agent reaches the files and records, beside the agent, what the task asks for and which opened file each item was closed against.

\paragraph{Agent Memory}
Memory for LLM agents is typically written by the model itself and kept across sessions or trials~\citep{MemorySurvey}: content is moved out of the context window into external storage and brought back when needed~\citep{MemGPT}, or the agent records its experiences and reflections and retrieves them later~\citep{GenerativeAgents, Reflexion, AMem}. Beyond memory that outlives a task, the model likewise keeps its own state within a single task: a plan written into its context~\citep{PlanAndSolve}, notes it writes on what it reads~\citep{SelfNotes}, a compact internal state it rewrites at each turn~\citep{MEM1}, or, in multi-agent systems such as Magentic-One~\citep{MagenticOne}, a plan and a progress record that an orchestrator LLM writes and judges. In recent work, such a record is kept for the agent and updated as the task proceeds, by the model itself or by further models that supervise it~\citep{LiveLedger, ProactiveMemoryAgent, LongHorizonHarness}. However, wherever these systems keep a record of the task, it is kept by a model, which writes it and marks its items complete. In contrast, in \modelname, the tab is kept by the environment: the environment writes it from what the agent read and listed (the agent adds only its requirements), its items are marked done only against a captured excerpt or set aside with a reason, and, kept outside the context window, it is not displaced and is returned to the agent once, when it tries to finish.

\section{Method}
\label{sec:method}

In this section, we first formalize direct workspace interaction, a setting in which the agent is given access to every file of a workspace but no persistent record of the state of its task, and then present \modelname, which supplies this record in the form of an environment-side tab kept beside the agent.

\subsection{Direct Workspace Interaction}
\label{sec:dwi}

\paragraph{Workspace Tasks}
Let $\mathcal{W}$ denote a workspace, the collection of files an organization already holds (such as reports, spreadsheets, slide decks, and notes), and let $t$ denote a task that asks for a new deliverable $\mathcal{D}$ (for instance, a report or a spreadsheet) to be produced from them. What characterizes such a task is that what it asks for rarely sits in one place: $t$ decomposes into many individual elements (a fact to look up, a figure to extract, a section to write) whose sources are scattered across the files of $\mathcal{W}$, and the deliverable is complete only when every one of them has been carried into $\mathcal{D}$. The task is carried out by an agent, an LLM typically equipped with terminal tools (such as $\texttt{read}$, which opens a file, and $\texttt{bash}$, which runs a shell command), which works in turns: in each turn, the model issues tool calls and receives their observations, writing $\mathcal{D}$ into an output folder of $\mathcal{W}$, and it finishes when it responds without a tool call, which we refer to as a stop.

\paragraph{Direct Workspace Interaction}
To reach the files, the agent follows direct corpus interaction~\citep{DCI}: instead of querying a retriever over an index, it applies its terminal tools directly to the raw files, so that any file in $\mathcal{W}$ can be listed, searched, and opened as it is. We define \emph{direct workspace interaction} (DWI) as the setting in which a workspace task is carried out with this form of access: given $\mathcal{W}$ and $t$, the agent interacts directly with the files of $\mathcal{W}$ over many turns and hands in $\mathcal{D}$ when it finishes. DWI thus inherits from direct corpus interaction how files are reached, but differs from it in what is asked of the agent: there, the output is a short answer that a few passages typically suffice to support; here, it is a deliverable that brings together content from many files, so that what the agent finds should remain available until it is written into $\mathcal{D}$.

\paragraph{The Missing Record of the Task}
In its plain form, however, DWI leaves this to the context window, where what the agent has found survives only as a passage in a growing stream of observations. To measure how often content shown to the agent is missing from the deliverable, we analyze DCI~\citep{DCI}, the plain form of DWI, with GPT-5.4 nano~\citep{GPT54} on Workspace-Bench~\citep{WorkspaceBench}, which checks each deliverable against rubrics, over three runs of its 100 tasks. We assign each rubric that DCI fails to a failure type with LLM annotators (agreeing with human labels at a Cohen's kappa of 0.83), and check whether each numeric value a rubric expects appears in the observations of the agent and in its deliverable.

\begin{wrapfigure}[14]{r}{0.40\textwidth}
    \centering
    \vspace{-0.16in}
    \includegraphics[width=\linewidth]{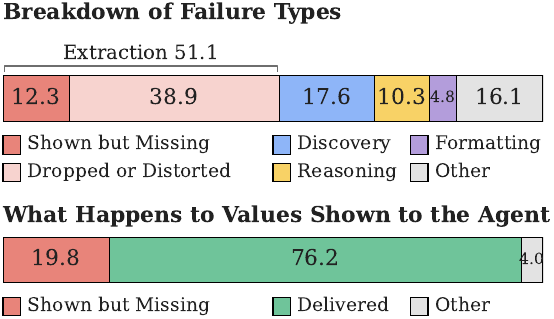}
    \vspace{-0.22in}
    \caption{DCI with GPT-5.4 nano on Workspace-Bench, as shares (\%). (Top) Lost points of the rubric pass rate by failure type. (Bottom) Rubric values shown to the agent, by outcome.}
    \label{fig:diagnosis}
    \vspace{-0.05in}
\end{wrapfigure}

As shown in \cref{fig:diagnosis} (top), extraction, in which the agent opened the right files but the deliverable lacks or distorts their content, accounts for 51.1\% of the points that DCI loses and splits into two parts: Shown but Missing (12.3\%), where the content is absent from the deliverable although it appears in the output of a file the agent had opened, and Dropped or Distorted (38.9\%), where the content is either absent with no such evidence or present but inexact, misplaced, or misread. Discovery, in which the right files are never opened (including files listed but not opened), accounts for another 17.6\%. Among the attempts that produced a deliverable, about one in five (19.8\%) of the 891 distinctive numeric values that the rubrics expect and that appear in an observation of the agent are absent from the deliverable, and the rubrics that expect them are not satisfied (\cref{fig:diagnosis}, bottom). By contrast, a record of the state of the task, kept outside the context window, offers what the stream cannot: at any point, it shows which elements of $t$ are still unmet, which observations have been used in $\mathcal{D}$, and which listed files have yet to be opened.

\begin{figure}[t!]
    \centering
    \includegraphics[width=\linewidth]{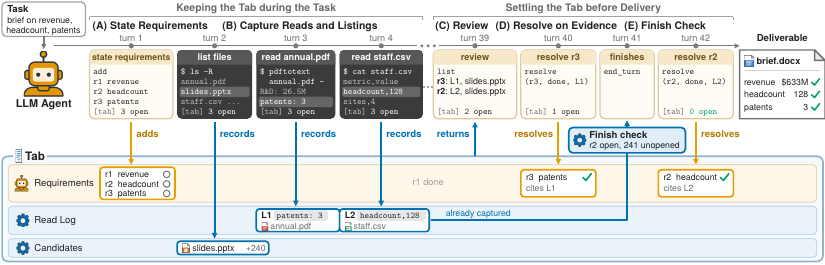}
    \vspace{-0.20in}
    \caption{\modelname over one task. The agent (top) works in turns, and the environment-side tab (bottom) fills under the turn that wrote each entry. (A) The agent states the requirements. (B) As the agent lists and reads files, the environment records each listed file as a candidate and each read as a read log entry with its anchor. (C) The review shows each open requirement beside its best read log match and the unopened files. (D) The agent resolves a requirement by citing the read log entry that holds its content. (E) At the stop, the environment returns one finish check with what is still open; here the tab had already captured the missing fact, so the agent closes it and delivers.}
    \label{fig:method}
    \vspace{-0.10in}
\end{figure}

Motivated by this, \modelname is built around an object that keeps such a record, the \emph{environment-side tab} (\cref{fig:method}), a per-task record of what the task still owes, maintained not by the model but by the environment (the loop that runs the model and its tool calls). Next, we define the tab (\cref{sec:tab}) and describe how it is kept during the task (\cref{sec:keeping}) and settled before delivery (\cref{sec:settling}).

\subsection{Environment-Side Tab}
\label{sec:tab}

We now turn to the environment-side tab (\cref{fig:method}), the central component of \modelname, and describe what it consists of, how the environment maintains it, and how the agent accesses it.

\paragraph{Composition of the Tab}
Recall that a deliverable is assembled from content scattered across many files, often long after that content was first encountered (\cref{sec:dwi}). Consequently, by the time the agent writes, it should rely on the following (remaining accessible): what the task asks for (so that no element is left out), what it has already read (so that content found earlier can be reused without searching for it again), and which files it has seen in a listing but not yet opened (so that it knows where to look for what is still missing). To this end, the tab maintains one kind of entry for each of them. Formally, for a task $t$, the tab is defined as $\mathcal{T}_t = (\mathcal{R}, \mathcal{L}, \mathcal{C})$, which consists of requirements ($\mathcal{R}$), a read log ($\mathcal{L}$), and candidates ($\mathcal{C}$), described as follows:
\begin{itemize}[leftmargin=*, itemsep=2pt, topsep=3pt]
    \item \textbf{Requirements ($\mathcal{R}$).} The requirements are the individual items that the task asks for, each being one fact, figure, or element of the deliverable. For instance, for a task that asks for a short report with several figures from an annual report, one requirement would be the number of new invention patents in 2024. At the start of the task, the agent states these items, and the environment records each of them as a requirement $r_i \in \mathcal{R}$ and tracks its status from then on. Specifically, a requirement is \texttt{open} when it is recorded, and the agent could later resolve it in one of two ways: as \texttt{done}, when its content has been found in a file, in which case the agent cites the entry of the read log that contains it; or as \texttt{unavailable}, when no file supplies it (for instance, because it is to be computed or written by the agent), in which case the agent states the reason.
    \item \textbf{Read log ($\mathcal{L}$).} The read log keeps a copy of what the agent has read. Whenever the agent reads a file (by opening it or extracting its text), the environment records an entry $\ell_j \in \mathcal{L}$, which consists of an excerpt (the content exactly as shown to the model) and an anchor (its provenance: the path of the file and the command that produced it). In the example above, extracting the text of the annual report yields an entry whose excerpt contains the line with the patent count, so it can be found again without reopening the report.
    \item \textbf{Candidates ($\mathcal{C}$).} The candidates are the files the agent has come across but not yet looked into. Whenever a file appears in a listing (such as the output of a command that lists a folder or searches for file names) without having been opened, the environment records it in $\mathcal{C} \subseteq \mathcal{W}$, and removes it once the agent opens it. In the example, if the folder of the annual report also contains a presentation of the same year that the agent has not opened, it remains among the candidates, as a place where a missing figure may still be found.
\end{itemize}
Note that it is the environment, rather than the model itself, that keeps every entry of the tab: it stores the requirements as the agent states them, and it captures the read log and the candidates on its own, since it observes every read and every listing, whereas a record that the model keeps for itself may miss some of them. Moreover, the tab is stored outside the context window, so that its entries are not displaced by new observations, and outside the workspace, so that it is not mixed with the files the agent lists and reads or with the deliverable. In other words, being kept by the environment and stored outside the context window is what distinguishes the tab; a record that the model keeps on its own within its context has neither of these properties.

\paragraph{Operations on the Tab}
Since the tab is stored outside the context window, the agent accesses it through four operations. Specifically, two of them update the requirements: $\texttt{add}$ states one or more requirements, and $\texttt{resolve}$ resolves a requirement as \texttt{done} or \texttt{unavailable}. The other two read the tab: $\texttt{list}$ lists the requirements with their status, the candidates, or the entries of the read log, and $\texttt{show}$ returns a single entry of the read log (its excerpt with its anchor), so that content read earlier can be consulted again without reopening the file. We describe when and how these operations are used in the following subsections.

\subsection{Keeping the Tab during the Task}
\label{sec:keeping}

We now describe how the tab is kept while the agent works. Since the tab is useful at a given turn only if it already holds what the agent needs then, each kind of entry is recorded as it arises, by the party in a position to record it: the agent states the requirements at the start, while the environment captures reads and listings from the observations and reports the state of the tab alongside them.

\paragraph{Stating the Requirements}
Of the three kinds of entries, the requirements are the one the environment cannot capture on its own, since what the task asks for appears only in the task, and dividing it into items takes an understanding of the task that the model supplies. \modelname therefore adds a short block to the system prompt, which asks the agent to begin by stating the requirements with $\texttt{add}$ and then to work as usual.

\paragraph{Capturing Reads and Listings}
In contrast, the read log and the candidates need no action from the agent: the environment sees every observation, so it records both on its own. For each observation it decides which files the agent has read and which it has only seen listed. However, the terminal itself draws no such line: many commands read a file, and a single command can show the content of one file with the names of many others. \modelname therefore draws this line from the command and its observation: a file counts as read when its content appears, and as listed when only its path does. An entry of the read log stores the observation as it is, neither parsed nor summarized, so that capturing depends on no file format and the entry holds what the agent saw. Since one listing can name hundreds of files while a candidate carries only its path, the candidates are ranked by the lexical relevance of their paths to the task with BM25~\citep{BM25}.

\paragraph{Reporting the State of the Tab}
Storing the tab outside the context window keeps its entries from being displaced, but the agent then sees none of them unless it calls an operation. The environment therefore appends to each observation a status line carrying the state of the tab: which requirements remain \texttt{open}, how much the read log and the candidates hold, and how long since its last operation.

\subsection{Settling the Tab before Delivery}
\label{sec:settling}

We now describe how the tab is settled before delivery. A requirement stays \texttt{open} until the agent resolves it, so a tab is settled once all its requirements are resolved. \modelname works toward that state with three mechanisms: a review that shows what may supply each open requirement; a resolution that names where the content of a requirement was found, or why no file supplies it; and a finish check sent at the stop.

\paragraph{Reviewing the Requirements}
Connecting a requirement to the file that supplies it is a link the context window leaves implicit, whereas the tab holds both ends of it: the requirements on one side, the read log and the candidates on the other. \modelname therefore offers a review that joins them: when the agent lists its requirements with $\texttt{list}$, the tab returns each open requirement beside the read-log entries that best match it and the unopened files still to try, both ranked by BM25 against the requirement. The prompt asks the agent to do so after exploring and before resolving any requirement, so that it can still act on what the review shows.

\paragraph{Resolving on Evidence}
To resolve a requirement as \texttt{done}, the agent cites the read-log entry its content came from, and the tab checks the citation so that a \texttt{done} rests on what the agent read: $\texttt{resolve}$ refuses a citation whose path and excerpt share no term with the requirement, while \texttt{unavailable} requires a reason.

\paragraph{Checking at the Stop}
To give the agent a final pass over the tab, \modelname places a check at the stop, the last moment to change the deliverable. At the first stop, the finish check reports the status of the requirements and the files listed but not opened, and asks the agent to add what is missing before finishing again.

\section{Experimental Setup}
\label{sec:setup}

\begin{table*}[t]
    \centering
    \caption{
    Main results on three benchmarks with three LLMs, averaged over three runs (standard deviation in parentheses). Best results within each LLM are \textbf{bolded}; second best are \underline{underlined}.
    }
    \label{tab:main}
    \vspace{-0.075in}
    \small
    \resizebox{\textwidth}{!}{
    \setlength{\tabcolsep}{5pt}
    \renewcommand{\arraystretch}{0.95}
    \begin{tabular}{l cc cc cc}
        \toprule
        & \multicolumn{2}{c}{\textbf{Workspace-Bench}} & \multicolumn{2}{c}{\textbf{TheAgentCompany}} & \multicolumn{2}{c}{\textbf{OfficeQA Pro}} \\
        \cmidrule[0.4pt](lr){2-3} \cmidrule[0.4pt](lr){4-5} \cmidrule[0.4pt](lr){6-7}
        \textbf{Method} & Pass Rate & TCR@70 & Score & Success & Acc@0\% & Acc@1\% \\

        \midrule

        \multicolumn{7}{c}{\cellcolor{mygray}\rule{0pt}{2.0ex}\textbf{\textit{GPT-5.4 nano}}} \\
        \addlinespace[1pt]
        \midrule
        DCI & 41.63\std{0.90} & 20.67\std{1.70} & 32.09\std{2.22} & 17.86\std{5.05} & 53.53\std{3.27} & 65.38\std{2.83} \\
        TODO & 41.67\std{0.95} & \underline{23.33}\std{2.62} & \underline{36.74}\std{2.74} & \underline{22.62}\std{3.37} & 54.81\std{3.14} & 65.71\std{3.27} \\
        Self-Refine & 41.32\std{2.15} & 22.33\std{3.09} & 35.76\std{2.97} & \underline{22.62}\std{4.45} & 54.17\std{2.40} & 64.74\std{2.40} \\
        Self-Tracking & \underline{42.93}\std{0.44} & 23.00\std{1.41} & 34.30\std{1.52} & 20.24\std{1.68} & \underline{55.13}\std{2.97} & \underline{66.99}\std{3.17} \\
        \rowcolor{citeblue!10}
        \textbf{\modelname (Ours)} & \textbf{44.73}\std{0.56} & \textbf{25.67}\std{2.49} & \textbf{39.12}\std{1.22} & \textbf{23.81}\std{1.68} & \textbf{60.58}\std{0.79} & \textbf{71.15}\std{0.79} \\
        \midrule
        \addlinespace[2pt]

        \multicolumn{7}{c}{\cellcolor{mygray}\rule{0pt}{2.0ex}\textbf{\textit{DeepSeek V4 Flash}}} \\
        \addlinespace[1pt]
        \midrule
        DCI & 53.28\std{1.29} & \underline{38.67}\std{1.25} & 37.47\std{1.66} & 21.43\std{2.92} & 65.71\std{1.63} & \underline{74.68}\std{1.20} \\
        TODO & \underline{54.94}\std{2.44} & 38.00\std{2.16} & \underline{40.34}\std{3.21} & 25.00\std{5.05} & \underline{66.03}\std{1.63} & 73.08\std{2.83} \\
        Self-Refine & 53.44\std{0.53} & 37.67\std{0.47} & 40.24\std{1.76} & \underline{26.19}\std{1.68} & 64.74\std{1.63} & 73.08\std{0.79} \\
        Self-Tracking & 53.56\std{1.57} & 37.33\std{3.30} & 39.24\std{5.21} & \underline{26.19}\std{6.07} & 64.74\std{1.63} & 73.08\std{0.79} \\
        \rowcolor{citeblue!10}
        \textbf{\modelname (Ours)} & \textbf{59.99}\std{0.60} & \textbf{41.00}\std{1.63} & \textbf{43.11}\std{0.61} & \textbf{29.76}\std{1.68} & \textbf{68.59}\std{1.20} & \textbf{77.56}\std{1.20} \\
        \midrule
        \addlinespace[2pt]

        \multicolumn{7}{c}{\cellcolor{mygray}\rule{0pt}{2.0ex}\textbf{\textit{Gemini 3.8 Flash}}} \\
        \addlinespace[1pt]
        \midrule
        DCI & 31.42\std{2.89} & 23.33\std{1.70} & 26.35\std{2.50} & \underline{19.05}\std{1.68} & 43.08\std{0.79} & 45.64\std{0.91} \\
        TODO & 27.52\std{5.62} & 22.33\std{3.30} & 20.97\std{5.05} & 14.29\std{5.05} & 44.04\std{0.79} & 47.24\std{0.45} \\
        Self-Refine & \underline{32.73}\std{2.92} & \underline{28.00}\std{2.94} & \underline{26.68}\std{6.54} & 16.67\std{3.37} & 41.79\std{1.20} & 45.64\std{0.45} \\
        Self-Tracking & 26.87\std{1.23} & 21.00\std{2.16} & 21.68\std{6.23} & 14.29\std{5.83} & \underline{45.00}\std{2.72} & \underline{48.21}\std{2.52} \\
        \rowcolor{citeblue!10}
        \textbf{\modelname (Ours)} & \textbf{46.91}\std{1.18} & \textbf{36.33}\std{3.30} & \textbf{32.93}\std{0.41} & \textbf{22.62}\std{1.68} & \textbf{55.77}\std{0.79} & \textbf{62.82}\std{1.98} \\
        \bottomrule
    \end{tabular}
    }
\end{table*}

In this section, we describe the benchmarks, the baselines, and the implementation details.

\vspace{-0.05in}
\paragraph{Benchmarks}
Direct workspace interaction targets tasks whose deliverable draws on many files of a workspace. We therefore use two benchmarks that pose such tasks and check the deliverable element by element (by rubrics or by checkpoints), and one benchmark of QA over enterprise documents:
\begin{itemize}[leftmargin=*, itemsep=3pt, parsep=0pt, topsep=-2pt]
    \item \textbf{Workspace-Bench}~\citep{WorkspaceBench} simulates the workspaces of five professional roles (such as an operations manager and a researcher), each holding thousands of heterogeneous files, and asks for deliverables that draw on files scattered across a workspace; it has 100 tasks and 1,850 evaluation instances (one per rubric), on which we report the Rubric Pass Rate (the share of satisfied rubrics) and the Task Completion Rate (TCR@70, the share of tasks with at least 70\% of their rubrics satisfied), following its protocol.
    \item \textbf{TheAgentCompany}~\citep{TheAgentCompany} simulates a software company whose tasks are checked by programmatic evaluators at checkpoints; we use a subset of its tasks that need no service other than the file server of the company (such as other websites or external models), with 88 evaluation instances (one per checkpoint), and report Success (the share of tasks whose checkpoints all pass) and Score (which gives each task half credit for the fraction of checkpoint points it earns and half for its success), as proposed by its authors.
    \item \textbf{OfficeQA Pro}~\citep{OfficeQAPro} lets us check whether \modelname also helps in the conventional setting of question answering, where the deliverable is a single answer: its workspace holds 697 files (U.S. Treasury Bulletins spanning nearly a century, about 89,000 pages), and the answer to each question is found, and often computed, from these files. We use a subset of its questions that need no external data source (such as exchange rates or price indices), with 104 evaluation instances (one per question), and report the accuracy at an allowable absolute relative error of 0\% and 1\% (Acc@0\% and Acc@1\%), as in its protocol.
\end{itemize}

\vspace{-0.05in}
\paragraph{Baselines and Our Model}
We compare \modelname against DCI and against three baselines in which the model, rather than an environment-side tab, keeps track of what the task asks for:
\begin{itemize}[leftmargin=*, itemsep=3pt, parsep=0pt, topsep=-2pt]
    \item \textbf{DCI}~\citep{DCI} is direct corpus interaction applied to the workspace: the agent works on the files with terminal tools, leaving the state of the task to its context window.
    \item \textbf{TODO} follows plan-first prompting~\citep{PlanAndSolve}, asking the agent to begin by writing a TODO list of every fact, figure, and deliverable element the task asks for.
    \item \textbf{Self-Refine}~\citep{SelfRefine} has the model review and refine its own output; at the first stop, one message asks the agent to re-check each of these elements against the source files it read and to make sure that each is in the deliverable before finishing again.
    \item \textbf{Self-Tracking} combines the two, as in model-kept records of task progress~\citep{MagenticOne, LiveLedger}: the agent lists these elements at the start and re-checks them at the end.
    \item \textbf{\modelname} is our framework, in which the environment-side tab keeps track of the task.
\end{itemize}

\vspace{-0.05in}
\paragraph{Implementation Details}
Following \citet{DCI}, all methods use DCI-Agent-Lite, a terminal agent built on the Pi harness\footnote{\url{https://github.com/earendil-works/pi}}, with its terminal tools (such as $\texttt{read}$ and $\texttt{bash}$), a budget of 300 turns, and runtime context management at level 3 (which truncates each observation and clears older observations as they accumulate), to which we add a time limit of one hour per task. We instantiate every method with three LLMs: GPT-5.4 nano~\citep{GPT54}, DeepSeek V4 Flash~\citep{DeepSeekV4}, and Gemini 3.8 Flash~\citep{Gemini38}. Where the evaluation calls for an LLM judge, we follow the protocol of each benchmark and use GPT-5.4 mini. For \modelname, the calls to the tab operations are removed from the trajectory that the evaluators read, and its settings stay the same across benchmarks and models. We report the average over three runs.

\section{Experimental Results}
\label{sec:results}

\paragraph{Main Results}
We report the main results in \cref{tab:main}, where \modelname achieves the best performance on every benchmark, metric, and LLM. In contrast, the baselines in which the model keeps track of the task (by listing what the task asks for at the start, re-checking the deliverable at the end, or both) remain far less effective, suggesting that the gains of \modelname come not from instructing the model to keep track of the task but from the record that the environment keeps on its behalf, outside the context window. These gains hold across workspace settings, from assembling deliverables from office files (Workspace-Bench) and carrying out company tasks (TheAgentCompany) to answering questions from decades of archives (OfficeQA Pro).

\newsavebox{\tabHoldsBox}
\newsavebox{\tabOpsBox}
\begin{table}[t]
    \centering
    \begin{lrbox}{\tabHoldsBox}
        \small
        \setlength{\tabcolsep}{3.5pt}
        \renewcommand{\arraystretch}{1.1}
        \begin{tabular}{@{}l ccc@{}}
            \toprule
            & GPT & DeepSeek & Gemini \\
            \midrule
            Requirements ($\mathcal{R}$) & 4 & 5 & 4 \\
            \quad resolved as \texttt{done} (\%) & 74.0 & 77.7 & 50.7 \\
            \quad reviewed beside a matching excerpt (\%) & 87.1 & 88.0 & 69.9 \\
            \noalign{\vskip 0.25ex}\cdashline{1-4}\noalign{\vskip 0.75ex}
            Read-log entries ($\mathcal{L}$) & 14 & 18 & 6 \\
            \quad distinct files & 5 & 11 & 5 \\
            \quad excerpted text (K characters) & 17.7 & 40.2 & 12.1 \\
            \noalign{\vskip 0.25ex}\cdashline{1-4}\noalign{\vskip 0.75ex}
            Candidates ($\mathcal{C}$), not opened & 147 & 166 & 294 \\
            \bottomrule
        \end{tabular}
    \end{lrbox}
    \begin{lrbox}{\tabOpsBox}
        \small
        \setlength{\tabcolsep}{3.5pt}
        \renewcommand{\arraystretch}{1.1}
        \begin{tabular}{@{}l l ccc@{}}
            \toprule
            \textbf{Operation} & \textbf{The agent} & GPT & DeepSeek & Gemini \\
            \midrule
            $\texttt{add}$ & states requirements before the check & 56.9 & 100.0 & 100.0 \\
            $\texttt{list}$ & pulls the review before the check & 24.7 & 27.2 & 100.0 \\
            $\texttt{resolve}$ & cites a reviewed entry for \texttt{done} & 25.1 & 21.6 & 76.9 \\
            $\texttt{show}$ & re-reads a stored entry & 74.6 & 32.4 & 8.8 \\
            \noalign{\vskip 0.25ex}\cdashline{1-5}\noalign{\vskip 0.75ex}
            Finish check & operates the tab after it & 72.6 & 8.7 & 5.1 \\
            & edits the deliverable after it & 17.1 & 5.2 & 0.5 \\
            \noalign{\vskip 0.25ex}\cdashline{1-5}\noalign{\vskip 0.75ex}
            Candidates & opens a listed candidate & 22.7 & 54.7 & 19.0 \\
            \bottomrule
        \end{tabular}
    \end{lrbox}
    \edef\tabScale{\fpeval{min(1, 0.965*\the\linewidth/(\the\wd\tabHoldsBox + \the\wd\tabOpsBox))}}%
    \typeout{usage tables: scale \tabScale, heights \the\dimexpr\ht\tabHoldsBox+\dp\tabHoldsBox\relax\space and \the\dimexpr\ht\tabOpsBox+\dp\tabOpsBox\relax}%
    \begin{minipage}[t]{\tabScale\wd\tabHoldsBox}
        \caption{What the tab holds per attempt on Workspace-Bench, as medians or, where marked, as shares (\%).}
        \label{tab:tab_holds}
    \end{minipage}
    \hfill
    \begin{minipage}[t]{\tabScale\wd\tabOpsBox}
        \caption{How the agent operates the tab: Workspace-Bench attempts (\%) in which it does what each row states.}
        \label{tab:tab_ops}
    \end{minipage}
    \par\vspace{-0.075in}
    \begin{minipage}[t]{\tabScale\wd\tabHoldsBox}
        \centering
        \scalebox{\tabScale}{\usebox{\tabHoldsBox}}
    \end{minipage}
    \hfill
    \begin{minipage}[t]{\tabScale\wd\tabOpsBox}
        \centering
        \scalebox{\tabScale}{\usebox{\tabOpsBox}}
    \end{minipage}
\end{table}

\paragraph{What the Tab Holds}
To understand what the environment-side tab provides to the agent, we first examine its contents on Workspace-Bench. As shown in \cref{tab:tab_holds}, the tab holds a substantial record of each task: a median of four to five requirements, 6 to 18 read-log entries (each storing the text that the agent saw when it read a file, together with the path of that file) drawn from 5 to 11 distinct files and holding 12K to 40K characters of excerpted text, and 147 to 294 files that were listed but not opened. Moreover, the review brings this record to each requirement: when the agent pulls the review, 69.9\% to 88.0\% of its open requirements are shown beside a read-log entry whose excerpt contains a line with words of the requirement, so that, for most requirements, the review points the agent to a matching passage from a file it has already read.

\begin{wrapfigure}[15]{R}{0.36\textwidth}
    \centering
    \vspace{-0.19in}
    \includegraphics[width=\linewidth]{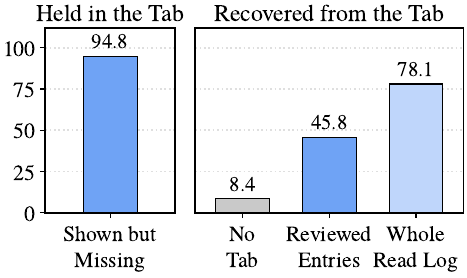}
    \vspace{-0.28in}
    \caption{\modelname with GPT-5.4 nano on Workspace-Bench (\%). (Left) Values held in the tab, among those the agent was shown but its deliverable lacks. (Right) Held values that an LLM recovers when they are masked, by the part of the tab given.}
    \label{fig:recovery}
    \vspace{-0.05in}
\end{wrapfigure}

\paragraph{Recovering Shown but Missing Content}
With DCI, about one in five of the rubric values shown to the agent is absent from its deliverable (\cref{fig:diagnosis}, bottom). We check whether, with \modelname, such values are held in the tab: for each numeric value that a failed rubric expects, that appears in an observation of the agent, and that is absent from the deliverable, we test whether the read log stores it. As shown in \cref{fig:recovery} (left), with GPT-5.4 nano, the tab holds 94.8\% of these values\footnote{The read log keeps only file content (\cref{sec:keeping}), not a value the agent computes (such as a sum) or sees only in a listing.}, so the missing content has in most cases been captured by the environment. We further test whether it is usable: for the requirements whose values the deliverable left out entirely, the same LLM, given the task and the requirement with its values masked, recovers 8.4\% of the values held in the tab without the tab, 45.8\% from the review of the tab for that requirement and the entries it chooses to open, and 78.1\% from the whole read log (\cref{fig:recovery}, right), so the review brings back more than half of what the full log provides while using a small fraction of its text. Consistently, \modelname succeeds on more rubrics in the Shown but Missing category (\cref{fig:diagnosis}, top) than DCI (27.0\% against 20.7\%).

\suppressfloats[t]
\begin{figure}[t]
    \centering
    \begin{minipage}[t]{0.44\textwidth}
        \vspace{0pt}
        \centering
    \captionof{table}{Ablation of \modelname (GPT-5.4 nano, Workspace-Bench, three runs), with drops in red.}
    \label{tab:ablation}
    \vspace{-0.05in}
    \small
    \resizebox{\linewidth}{!}{
    \setlength{\tabcolsep}{3.5pt}
    \renewcommand{\arraystretch}{1.1}
    \begin{tabular}{l ll}
        \toprule
        \textbf{Method} & Pass Rate & TCR@70 \\
        \midrule
        \rowcolor{citeblue!10}
        \textbf{\modelname (Ours)} & \textbf{44.73} & \textbf{25.67} \\
        \quad No Environmental Capture & 41.36\negdelta{3.37} & 22.33\negdelta{3.34} \\
        \quad No Review & 41.87\negdelta{2.86} & 20.33\negdelta{5.34} \\
        \quad No Evidence Check & 42.89\negdelta{1.84} & 24.33\negdelta{1.34} \\
        \quad No Finish Check & 42.89\negdelta{1.84} & 23.00\negdelta{2.67} \\
        \noalign{\vskip 0.25ex}\cdashline{1-3}\noalign{\vskip 0.75ex}
        DCI & 41.63 & 20.67 \\
        \bottomrule
    \end{tabular}
    }

    \end{minipage}
    \hfill
    \begin{minipage}[t]{0.53\textwidth}
        \vspace{0pt}
        \input{figures/reading}
    \end{minipage}
    \vspace{-0.1in}
\end{figure}

\paragraph{How the Agent Operates the Tab}
We then examine how the agent operates the tab, reported in \cref{tab:tab_ops}, which shows that each LLM draws on the tab in its own way. Specifically, Gemini 3.8 Flash works mainly through the review: it states its requirements and pulls the review before the finish check in every analyzed attempt, and in 76.9\% of attempts it resolves a requirement as \texttt{done} by citing a read-log entry that the review listed for that requirement. GPT-5.4 nano instead relies on the finish check: it often turns to the tab only when the check prompts it, and after the check it operates the tab in 72.6\% of attempts and edits its deliverable in 17.1\%; it also re-reads stored entries with $\texttt{show}$ in 74.6\% of attempts. Lastly, DeepSeek V4 Flash works mainly through the candidates: like Gemini 3.8 Flash, it states its requirements before the finish check in every analyzed attempt, but it then explores more widely, reading the most distinct files (a median of 11 in \cref{tab:tab_holds}) and opening a file that the tab listed as not yet opened in 54.7\% of attempts.

\paragraph{Ablation Studies}
To examine how much each component of the environment-side tab contributes to the performance gains, we remove one component at a time on Workspace-Bench with GPT-5.4 nano and report the results in \cref{tab:ablation}. From this, we observe that removing any of the four components lowers both the Pass Rate and TCR@70, indicating that each of them plays a role in these gains. Among them, removing the environmental capture (in which the environment no longer records the read log and the candidates, so that the review and the evidence check, which draw on them, are removed as well) leads to the largest drop in Pass Rate, down to the level of DCI (41.36 against 41.63), which indicates that the gains of \modelname come largely from what the environment captures on its own, namely the files that the agent reads and lists. Following this, removing the review leads to the next largest drop, which suggests that the captured excerpts are most useful when the agent sees them beside the requirements they match. Lastly, the evidence check and the finish check complement the capture and the review at two moments of the task: when the agent resolves a requirement as \texttt{done}, the evidence check keeps that resolution grounded in what the agent read, and when the agent stops, the finish check gives it a final pass over the tab before delivery.

\paragraph{Gains on Reading-Heavy Tasks}
Since the tab keeps what the agent has read outside the context window, its benefit is expected to grow with how much the agent reads. To see whether this holds, we split the tasks of Workspace-Bench into thirds by reading volume (the characters that the agent receives from its reads and shell commands). As shown in \cref{fig:reading}, the gain of \modelname over DCI generally grows with reading volume across the three LLMs, which indicates that the tab is most valuable when the agent has the most to keep track of (on average, the gain on the third of tasks with the most reading is twice that on the third with the least). Also, it is worth noting that the reading-heavy tasks are where DCI performs worst (its Pass Rate drops by 16 to 27 points from the lightest to the heaviest third); this suggests that the content read earlier remains in the context yet goes unused as observations accumulate, which is precisely what the tab keeps within reach.

\section{Conclusion}
\label{sec:conclusion}

In this work, we presented \modelname, a framework for direct workspace interaction, the setting in which an agent searches and reads the files of a workspace from a terminal to produce a deliverable. Motivated by the observation that what the task asks for and what the agent has read are typically left to the context window, we equipped the agent with an environment-side tab: a per-task record that the environment keeps outside the context window. Specifically, the agent states the requirements, and the environment records each read as an excerpt with its source and each file listed but not opened as a candidate. The agent can then review each open requirement beside its best-matching excerpts and candidates, resolve it by citing the excerpt that supplies it or by giving a reason, and receive a finish check at its stop. Empirical evaluations on three benchmarks with three LLMs show that \modelname improves over direct workspace interaction alone and over baselines in which the model itself keeps track of the task. Ablations attribute most of this gain to what the environment captures on its own, and the tab usually holds the values a deliverable needs once seen. We believe these findings position the environment-side tab as a natural complement to direct access: terminal tools let the agent reach any file, and the tab keeps track of the task until the deliverable is handed in.

\section*{Ethics Statement}

This work aims to improve how LLM agents produce deliverables from the files of a workspace, using existing models and public benchmarks. However, agents with \modelname may retain the biases and errors of their models, and since the tab stores excerpts of the files the agent reads, it may hold sensitive content in real workspaces. We therefore encourage safeguards, such as keeping the tab under the same access controls as the workspace and reviewing the deliverables, when deploying such agents in real-world settings.

\bibliographystyle{paper}
\bibliography{paper}

@inproceedings{TheAgentCompany,
  author       = {Frank Fangzheng Xu and
                  Yufan Song and
                  Boxuan Li and
                  Yuxuan Tang and
                  Kritanjali Jain and
                  Mengxue Bao and
                  Zora Zhiruo Wang and
                  Xuhui Zhou and
                  Zhitong Guo and
                  Murong Cao and
                  Mingyang Yang and
                  Hao Yang Lu and
                  Amaad Martin and
                  Zhe Su and
                  Leander Maben and
                  Raj Mehta and
                  Wayne Chi and
                  Lawrence Jang and
                  Yiqing Xie and
                  Shuyan Zhou and
                  Graham Neubig},
  editor       = {Danielle Belgrave and
                  Cheng Zhang and
                  Laura N. Montoya and
                  Hsuan{-}Tien Lin and
                  Razvan Pascanu and
                  Piotr Koniusz and
                  Marzyeh Ghassemi and
                  Nancy Chen and
                  Iv{\'{a}}n Vladimir Meza Ru{\'{\i}}z and
                  Arturo Loaiza{-}Bonilla},
  title        = {TheAgentCompany: Benchmarking {LLM} Agents on Consequential Real World
                  Tasks},
  booktitle    = {Advances in Neural Information Processing Systems 38: Annual Conference
                  on Neural Information Processing Systems 2025, NeurIPS 2025, San Diego,
                  CA, USA, December 2-7, 2025 / Mexico City, Mexico, November 30 - December
                  5, 2025},
  year         = {2025},
  url          = {http://papers.nips.cc/paper\_files/paper/2025/hash/0d744742f6fac4d1134c019b7cef3c8a-Abstract-Datasets\_and\_Benchmarks\_Track.html}
}

@article{GDPval,
  author       = {Tejal Patwardhan and
                  Rachel Dias and
                  Elizabeth Proehl and
                  Grace Kim and
                  Michele Wang and
                  Olivia Watkins and
                  Sim{\'{o}}n Posada Fishman and
                  Marwan Aljubeh and
                  Phoebe Thacker and
                  Laurance Fauconnet and
                  Natalie S. Kim and
                  Patrick Chao and
                  Samuel Miserendino and
                  Gildas Chabot and
                  David Li and
                  Michael Sharman and
                  Alexandra Barr and
                  Amelia Glaese and
                  Jerry Tworek},
  title        = {GDPval: Evaluating {AI} Model Performance on Real-World Economically
                  Valuable Tasks},
  journal      = {arXiv preprint arXiv:2510.04374},
  year         = {2025},
  url          = {https://doi.org/10.48550/arXiv.2510.04374}
}

@article{WorkspaceBench,
  author       = {Zirui Tang and
                  Xuanhe Zhou and
                  Yumou Liu and
                  Linchun Li and
                  Yukai Wu and
                  Weizheng Wang and
                  Hongzhang Huang and
                  Wei Zhou and
                  Jun Zhou and
                  Jiachen Song and
                  Shaoli Yu and
                  Jinqi Wang and
                  Zihang Zhou and
                  Hongyi Zhou and
                  Yuting Lv and
                  Jinyang Li and
                  Jiashuo Liu and
                  Ruoyu Chen and
                  Chunwei Liu and
                  Guoliang Li and
                  Jihua Kang and
                  Fan Wu},
  title        = {Workspace-Bench 1.0: Benchmarking {AI} Agents on Workspace Tasks with
                  Large-Scale File Dependencies},
  journal      = {arXiv preprint arXiv:2605.03596},
  year         = {2026},
  url          = {https://doi.org/10.48550/arXiv.2605.03596}
}

@inproceedings{ReAct,
  author       = {Shunyu Yao and
                  Jeffrey Zhao and
                  Dian Yu and
                  Nan Du and
                  Izhak Shafran and
                  Karthik R. Narasimhan and
                  Yuan Cao},
  title        = {ReAct: Synergizing Reasoning and Acting in Language Models},
  booktitle    = {The Eleventh International Conference on Learning Representations,
                  {ICLR} 2023, Kigali, Rwanda, May 1-5, 2023},
  publisher    = {OpenReview.net},
  year         = {2023},
  url          = {https://openreview.net/forum?id=WE\_vluYUL-X}
}

@inproceedings{Toolformer,
  author       = {Timo Schick and
                  Jane Dwivedi{-}Yu and
                  Roberto Dess{\`{\i}} and
                  Roberta Raileanu and
                  Maria Lomeli and
                  Eric Hambro and
                  Luke Zettlemoyer and
                  Nicola Cancedda and
                  Thomas Scialom},
  editor       = {Alice Oh and
                  Tristan Naumann and
                  Amir Globerson and
                  Kate Saenko and
                  Moritz Hardt and
                  Sergey Levine},
  title        = {Toolformer: Language Models Can Teach Themselves to Use Tools},
  booktitle    = {Advances in Neural Information Processing Systems 36: Annual Conference
                  on Neural Information Processing Systems 2023, NeurIPS 2023, New Orleans,
                  LA, USA, December 10 - 16, 2023},
  year         = {2023},
  url          = {http://papers.nips.cc/paper\_files/paper/2023/hash/d842425e4bf79ba039352da0f658a906-Abstract-Conference.html}
}

@inproceedings{SWEAgent,
  author       = {John Yang and
                  Carlos E. Jimenez and
                  Alexander Wettig and
                  Kilian Lieret and
                  Shunyu Yao and
                  Karthik Narasimhan and
                  Ofir Press},
  editor       = {Amir Globersons and
                  Lester Mackey and
                  Danielle Belgrave and
                  Angela Fan and
                  Ulrich Paquet and
                  Jakub M. Tomczak and
                  Cheng Zhang},
  title        = {SWE-agent: Agent-Computer Interfaces Enable Automated Software Engineering},
  booktitle    = {Advances in Neural Information Processing Systems 37: Annual Conference
                  on Neural Information Processing Systems 2024, NeurIPS 2024, Vancouver,
                  BC, Canada, December 10 - 15, 2024},
  year         = {2024},
  url          = {http://papers.nips.cc/paper\_files/paper/2024/hash/5a7c947568c1b1328ccc5230172e1e7c-Abstract-Conference.html}
}

@article{TerminalBench,
  author       = {Mike A. Merrill and
                  Alexander Glenn Shaw and
                  Nicholas Carlini and
                  Boxuan Li and
                  Harsh Raj and
                  Ivan Bercovich and
                  Lin Shi and
                  Jeong Yeon Shin and
                  Thomas Walshe and
                  Estefany Kelly Buchanan and
                  Junhong Shen and
                  Guanghao Ye and
                  Haowei Lin and
                  Jason Poulos and
                  Maoyu Wang and
                  Marianna Nezhurina and
                  Jenia Jitsev and
                  Di Lu and
                  Orfeas Menis{-}Mastromichalakis and
                  Zhiwei Xu and
                  Zizhao Chen and
                  Yue Liu and
                  Robert Zhang and
                  Leon Liangyu Chen and
                  Anurag Kashyap and
                  Jan{-}Lucas Uslu and
                  Jeffrey Li and
                  Jianbo Wu and
                  Minghao Yan and
                  Song Bian and
                  Vedang Sharma and
                  Ke Sun and
                  Steven Dillmann and
                  Akshay Anand and
                  Andrew Lanpouthakoun and
                  Bardia Koopah and
                  Changran Hu and
                  Etash Kumar Guha and
                  Gabriel H. S. Dreiman and
                  Jiacheng Zhu and
                  Karl Krauth and
                  Li Zhong and
                  Niklas Muennighoff and
                  Robert Amanfu and
                  Shangyin Tan and
                  Shreyas Pimpalgaonkar and
                  Tushar Aggarwal and
                  Xiangning Lin and
                  Xin Lan and
                  Xuandong Zhao and
                  Yiqing Liang and
                  Yuanli Wang and
                  Zilong Wang and
                  Changzhi Zhou and
                  David Heineman and
                  Hange Liu and
                  Harsh Trivedi and
                  John Yang and
                  Junhong Lin and
                  Manish Shetty and
                  Michael Yang and
                  Nabil Omi and
                  Negin Raoof and
                  Shanda Li and
                  Terry Yue Zhuo and
                  Wuwei Lin and
                  Yiwei Dai and
                  Yuxin Wang and
                  Wenhao Chai and
                  Shang Zhou and
                  Dariush Wahdany and
                  Ziyu She and
                  Jiaming Hu and
                  Zhikang Dong and
                  Yuxuan Zhu and
                  Sasha Cui and
                  Ahson Saiyed and
                  Arinbj{\"{o}}rn Kolbeinsson and
                  Jesse Hu and
                  Christopher Michael Rytting and
                  Ryan Marten and
                  Yixin Wang and
                  Alex Dimakis and
                  Andy Konwinski and
                  Ludwig Schmidt},
  title        = {Terminal-Bench: Benchmarking Agents on Hard, Realistic Tasks in Command
                  Line Interfaces},
  journal      = {arXiv preprint arXiv:2601.11868},
  year         = {2026},
  url          = {https://doi.org/10.48550/arXiv.2601.11868}
}

@article{DCI,
  author       = {Zhuofeng Li and
                  Haoxiang Zhang and
                  Cong Wei and
                  Pan Lu and
                  Ping Nie and
                  Yi Lu and
                  Yuyang Bai and
                  Shangbin Feng and
                  Hangxiao Zhu and
                  Ming Zhong and
                  Yuyu Zhang and
                  Jianwen Xie and
                  Yejin Choi and
                  James Zou and
                  Jiawei Han and
                  Wenhu Chen and
                  Jimmy Lin and
                  Dongfu Jiang and
                  Yu Zhang},
  title        = {Beyond Semantic Similarity: Rethinking Retrieval for Agentic Search
                  via Direct Corpus Interaction},
  journal      = {arXiv preprint arXiv:2605.05242},
  year         = {2026},
  url          = {https://doi.org/10.48550/arXiv.2605.05242}
}

@inproceedings{DPR,
  author       = {Vladimir Karpukhin and
                  Barlas Oguz and
                  Sewon Min and
                  Patrick Lewis and
                  Ledell Wu and
                  Sergey Edunov and
                  Danqi Chen and
                  Wen{-}tau Yih},
  editor       = {Bonnie Webber and
                  Trevor Cohn and
                  Yulan He and
                  Yang Liu},
  title        = {Dense Passage Retrieval for Open-Domain Question Answering},
  booktitle    = {Proceedings of the 2020 Conference on Empirical Methods in Natural
                  Language Processing, {EMNLP} 2020, Online, November 16-20, 2020},
  pages        = {6769--6781},
  publisher    = {Association for Computational Linguistics},
  year         = {2020},
  url          = {https://doi.org/10.18653/v1/2020.emnlp-main.550}
}

@inproceedings{RAG,
  author       = {Patrick Lewis and
                  Ethan Perez and
                  Aleksandra Piktus and
                  Fabio Petroni and
                  Vladimir Karpukhin and
                  Naman Goyal and
                  Heinrich K{\"{u}}ttler and
                  Mike Lewis and
                  Wen{-}tau Yih and
                  Tim Rockt{\"{a}}schel and
                  Sebastian Riedel and
                  Douwe Kiela},
  editor       = {Hugo Larochelle and
                  Marc'Aurelio Ranzato and
                  Raia Hadsell and
                  Maria{-}Florina Balcan and
                  Hsuan{-}Tien Lin},
  title        = {Retrieval-Augmented Generation for Knowledge-Intensive {NLP} Tasks},
  booktitle    = {Advances in Neural Information Processing Systems 33: Annual Conference
                  on Neural Information Processing Systems 2020, NeurIPS 2020, December
                  6-12, 2020, virtual},
  year         = {2020},
  url          = {https://proceedings.neurips.cc/paper/2020/hash/6b493230205f780e1bc26945df7481e5-Abstract.html}
}

@article{DRDCI,
  author       = {Yi Lu and
                  Zhuofeng Li and
                  Ping Nie and
                  Haoxiang Zhang and
                  Yuyu Zhang and
                  Kai Zou and
                  Wenhu Chen and
                  Jimmy Lin and
                  Dongfu Jiang and
                  Yu Zhang},
  title        = {Dr-DCI: Scaling Direct Corpus Interaction via Dynamic Workspace Expansion},
  journal      = {arXiv preprint arXiv:2606.14885},
  year         = {2026},
  url          = {https://doi.org/10.48550/arXiv.2606.14885}
}

@article{GrepSeek,
  author       = {Alireza Salemi and
                  Chang Zeng and
                  Atharva Nijasure and
                  Jui{-}Hui Chung and
                  Razieh Rahimi and
                  Fernando Diaz and
                  Hamed Zamani},
  title        = {GrepSeek: Training Search Agents for Direct Corpus Interaction},
  journal      = {arXiv preprint arXiv:2605.29307},
  year         = {2026},
  url          = {https://doi.org/10.48550/arXiv.2605.29307}
}

@article{IsGrepAllYouNeed,
  author       = {Sahil Sen and
                  Akhil Kasturi and
                  Elias Lumer and
                  Anmol Gulati and
                  Vamse Kumar Subbiah},
  title        = {Is Grep All You Need? How Agent Harnesses Reshape Agentic Search},
  journal      = {arXiv preprint arXiv:2605.15184},
  year         = {2026},
  url          = {https://doi.org/10.48550/arXiv.2605.15184}
}

@article{KeywordSearch,
  author       = {Shreyas Subramanian and
                  Adewale Akinfaderin and
                  Yanyan Zhang and
                  Ishan Singh and
                  Mani Khanuja and
                  Sandeep Singh and
                  Maira Ladeira Tanke},
  title        = {Keyword search is all you need: Achieving RAG-Level Performance without
                  vector databases using agentic tool use},
  journal      = {arXiv preprint arXiv:2602.23368},
  year         = {2026},
  url          = {https://doi.org/10.48550/arXiv.2602.23368}
}

@inproceedings{OpenHands,
  author       = {Xingyao Wang and
                  Boxuan Li and
                  Yufan Song and
                  Frank F. Xu and
                  Xiangru Tang and
                  Mingchen Zhuge and
                  Jiayi Pan and
                  Yueqi Song and
                  Bowen Li and
                  Jaskirat Singh and
                  Hoang H. Tran and
                  Fuqiang Li and
                  Ren Ma and
                  Mingzhang Zheng and
                  Bill Qian and
                  Yanjun Shao and
                  Niklas Muennighoff and
                  Yizhe Zhang and
                  Binyuan Hui and
                  Junyang Lin and
                  et al.},
  title        = {OpenHands: An Open Platform for {AI} Software Developers as Generalist
                  Agents},
  booktitle    = {The Thirteenth International Conference on Learning Representations,
                  {ICLR} 2025, Singapore, April 24-28, 2025},
  publisher    = {OpenReview.net},
  year         = {2025},
  url          = {https://openreview.net/forum?id=OJd3ayDDoF}
}

@inproceedings{OSWorld,
  author       = {Tianbao Xie and
                  Danyang Zhang and
                  Jixuan Chen and
                  Xiaochuan Li and
                  Siheng Zhao and
                  Ruisheng Cao and
                  Toh Jing Hua and
                  Zhoujun Cheng and
                  Dongchan Shin and
                  Fangyu Lei and
                  Yitao Liu and
                  Yiheng Xu and
                  Shuyan Zhou and
                  Silvio Savarese and
                  Caiming Xiong and
                  Victor Zhong and
                  Tao Yu},
  editor       = {Amir Globersons and
                  Lester Mackey and
                  Danielle Belgrave and
                  Angela Fan and
                  Ulrich Paquet and
                  Jakub M. Tomczak and
                  Cheng Zhang},
  title        = {OSWorld: Benchmarking Multimodal Agents for Open-Ended Tasks in Real
                  Computer Environments},
  booktitle    = {Advances in Neural Information Processing Systems 37: Annual Conference
                  on Neural Information Processing Systems 2024, NeurIPS 2024, Vancouver,
                  BC, Canada, December 10 - 15, 2024},
  year         = {2024},
  url          = {http://papers.nips.cc/paper\_files/paper/2024/hash/5d413e48f84dc61244b6be550f1cd8f5-Abstract-Datasets\_and\_Benchmarks\_Track.html}
}

@article{OfficeBench,
  author       = {Zilong Wang and
                  Yuedong Cui and
                  Li Zhong and
                  Zimin Zhang and
                  Da Yin and
                  Bill Yuchen Lin and
                  Jingbo Shang},
  title        = {OfficeBench: Benchmarking Language Agents across Multiple Applications
                  for Office Automation},
  journal      = {arXiv preprint arXiv:2407.19056},
  year         = {2024},
  url          = {https://doi.org/10.48550/arXiv.2407.19056}
}

@article{OfficeQAPro,
  author       = {Krista Opsahl{-}Ong and
                  Arnav Singhvi and
                  Jasmine Collins and
                  Ivan Zhou and
                  Cindy Wang and
                  Ashutosh Baheti and
                  Owen Oertell and
                  Jacob Portes and
                  Sam Havens and
                  Erich Elsen and
                  Michael Bendersky and
                  Matei Zaharia and
                  Xing Chen},
  title        = {OfficeQA Pro: An Enterprise Benchmark for End-to-End Grounded Reasoning},
  journal      = {arXiv preprint arXiv:2603.08655},
  year         = {2026},
  url          = {https://doi.org/10.48550/arXiv.2603.08655}
}

@article{TUABench,
  author       = {Shoufa Chen and
                  Luyuan Wang and
                  Xuan Yang and
                  Zhiheng Liu and
                  Yuren Cong and
                  Yuanfeng Ji and
                  Feiyan Zhou and
                  Xiaohui Zhang and
                  Fanny Yang and
                  Belinda Zeng},
  title        = {TUA-Bench: {A} Benchmark for General-Purpose Terminal-Use Agents},
  journal      = {arXiv preprint arXiv:2606.28480},
  year         = {2026},
  url          = {https://doi.org/10.48550/arXiv.2606.28480}
}

@article{SearchR1,
  author       = {Bowen Jin and
                  Hansi Zeng and
                  Zhenrui Yue and
                  Dong Wang and
                  Hamed Zamani and
                  Jiawei Han},
  title        = {Search-R1: Training LLMs to Reason and Leverage Search Engines with
                  Reinforcement Learning},
  journal      = {arXiv preprint arXiv:2503.09516},
  year         = {2025},
  url          = {https://doi.org/10.48550/arXiv.2503.09516}
}

@inproceedings{SearchO1,
  author       = {Xiaoxi Li and
                  Guanting Dong and
                  Jiajie Jin and
                  Yuyao Zhang and
                  Yujia Zhou and
                  Yutao Zhu and
                  Peitian Zhang and
                  Zhicheng Dou},
  editor       = {Christos Christodoulopoulos and
                  Tanmoy Chakraborty and
                  Carolyn Rose and
                  Violet Peng},
  title        = {Search-o1: Agentic Search-Enhanced Large Reasoning Models},
  booktitle    = {Proceedings of the 2025 Conference on Empirical Methods in Natural
                  Language Processing, {EMNLP} 2025, Suzhou, China, November 4-9, 2025},
  pages        = {5420--5438},
  publisher    = {Association for Computational Linguistics},
  year         = {2025},
  url          = {https://doi.org/10.18653/v1/2025.emnlp-main.276}
}

@article{InteractionSpaces,
  author       = {Shengyao Zhuang and
                  Yuansheng Ni and
                  Hengxin Fun and
                  Jimmy Lin and
                  Xueguang Ma},
  title        = {Towards Retrieving Interaction Spaces for Agentic Search},
  journal      = {arXiv preprint arXiv:2606.06880},
  year         = {2026},
  url          = {https://doi.org/10.48550/arXiv.2606.06880}
}

@article{RelevanceGuidingCorpusInteraction,
  author       = {Jiangnan Li and
                  Yuqing Li and
                  Mo Yu and
                  Jinchao Zhang and
                  Jie Zhou},
  title        = {A New Role for Relevance: Guiding Corpus Interaction in Agentic Search},
  journal      = {arXiv preprint arXiv:2607.24223},
  year         = {2026},
  url          = {https://doi.org/10.48550/arXiv.2607.24223}
}

@article{MemorySurvey,
  author       = {Zeyu Zhang and
                  Quanyu Dai and
                  Xiaohe Bo and
                  Chen Ma and
                  Rui Li and
                  Xu Chen and
                  Jieming Zhu and
                  Zhenhua Dong and
                  Ji{-}Rong Wen},
  title        = {A Survey on the Memory Mechanism of Large Language Model-based Agents},
  journal      = {{ACM} Trans. Inf. Syst.},
  volume       = {43},
  number       = {6},
  pages        = {155:1--155:47},
  year         = {2025},
  url          = {https://doi.org/10.1145/3748302}
}

@article{MemGPT,
  author       = {Charles Packer and
                  Vivian Fang and
                  Shishir G. Patil and
                  Kevin Lin and
                  Sarah Wooders and
                  Joseph E. Gonzalez},
  title        = {MemGPT: Towards LLMs as Operating Systems},
  journal      = {arXiv preprint arXiv:2310.08560},
  year         = {2023},
  url          = {https://doi.org/10.48550/arXiv.2310.08560}
}

@inproceedings{GenerativeAgents,
  author       = {Joon Sung Park and
                  Joseph C. O'Brien and
                  Carrie Jun Cai and
                  Meredith Ringel Morris and
                  Percy Liang and
                  Michael S. Bernstein},
  editor       = {Sean Follmer and
                  Jeff Han and
                  J{\"{u}}rgen Steimle and
                  Nathalie Henry Riche},
  title        = {Generative Agents: Interactive Simulacra of Human Behavior},
  booktitle    = {Proceedings of the 36th Annual {ACM} Symposium on User Interface Software
                  and Technology, {UIST} 2023, San Francisco, CA, USA, 29 October 2023-
                  1 November 2023},
  pages        = {2:1--2:22},
  publisher    = {{ACM}},
  year         = {2023},
  url          = {https://doi.org/10.1145/3586183.3606763}
}

@inproceedings{Reflexion,
  author       = {Noah Shinn and
                  Federico Cassano and
                  Ashwin Gopinath and
                  Karthik Narasimhan and
                  Shunyu Yao},
  editor       = {Alice Oh and
                  Tristan Naumann and
                  Amir Globerson and
                  Kate Saenko and
                  Moritz Hardt and
                  Sergey Levine},
  title        = {Reflexion: language agents with verbal reinforcement learning},
  booktitle    = {Advances in Neural Information Processing Systems 36: Annual Conference
                  on Neural Information Processing Systems 2023, NeurIPS 2023, New Orleans,
                  LA, USA, December 10 - 16, 2023},
  year         = {2023},
  url          = {http://papers.nips.cc/paper\_files/paper/2023/hash/1b44b878bb782e6954cd888628510e90-Abstract-Conference.html}
}

@inproceedings{AMem,
  author       = {Wujiang Xu and
                  Zujie Liang and
                  Kai Mei and
                  Hang Gao and
                  Juntao Tan and
                  Yongfeng Zhang},
  editor       = {Danielle Belgrave and
                  Cheng Zhang and
                  Laura N. Montoya and
                  Hsuan{-}Tien Lin and
                  Razvan Pascanu and
                  Piotr Koniusz and
                  Marzyeh Ghassemi and
                  Nancy Chen and
                  Iv{\'{a}}n Vladimir Meza Ru{\'{\i}}z and
                  Arturo Loaiza{-}Bonilla},
  title        = {A-Mem: Agentic Memory for {LLM} Agents},
  booktitle    = {Advances in Neural Information Processing Systems 38: Annual Conference
                  on Neural Information Processing Systems 2025, NeurIPS 2025, San Diego,
                  CA, USA, December 2-7, 2025 / Mexico City, Mexico, November 30 - December
                  5, 2025},
  year         = {2025},
  url          = {http://papers.nips.cc/paper\_files/paper/2025/hash/19909c36f51abc4856b4560aff3d36d6-Abstract-Conference.html}
}

@inproceedings{PlanAndSolve,
  author       = {Lei Wang and
                  Wanyu Xu and
                  Yihuai Lan and
                  Zhiqiang Hu and
                  Yunshi Lan and
                  Roy Ka{-}Wei Lee and
                  Ee{-}Peng Lim},
  editor       = {Anna Rogers and
                  Jordan L. Boyd{-}Graber and
                  Naoaki Okazaki},
  title        = {Plan-and-Solve Prompting: Improving Zero-Shot Chain-of-Thought Reasoning
                  by Large Language Models},
  booktitle    = {Proceedings of the 61st Annual Meeting of the Association for Computational
                  Linguistics (Volume 1: Long Papers), {ACL} 2023, Toronto, Canada,
                  July 9-14, 2023},
  pages        = {2609--2634},
  publisher    = {Association for Computational Linguistics},
  year         = {2023},
  url          = {https://doi.org/10.18653/v1/2023.acl-long.147}
}

@article{MEM1,
  author       = {Zijian Zhou and
                  Ao Qu and
                  Zhaoxuan Wu and
                  Sunghwan Kim and
                  Alok Prakash and
                  Daniela Rus and
                  Jinhua Zhao and
                  Bryan Kian Hsiang Low and
                  Paul Pu Liang},
  title        = {{MEM1:} Learning to Synergize Memory and Reasoning for Efficient Long-Horizon
                  Agents},
  journal      = {arXiv preprint arXiv:2506.15841},
  year         = {2025},
  url          = {https://doi.org/10.48550/arXiv.2506.15841}
}

@article{MagenticOne,
  author       = {Adam Fourney and
                  Gagan Bansal and
                  Hussein Mozannar and
                  Cheng Tan and
                  Eduardo Salinas and
                  Erkang Zhu and
                  Friederike Niedtner and
                  Grace Proebsting and
                  Griffin Bassman and
                  Jack Gerrits and
                  Jacob Alber and
                  Peter Chang and
                  Ricky Loynd and
                  Robert West and
                  Victor Dibia and
                  Ahmed Awadallah and
                  Ece Kamar and
                  Rafah Hosn and
                  Saleema Amershi},
  title        = {Magentic-One: {A} Generalist Multi-Agent System for Solving Complex
                  Tasks},
  journal      = {arXiv preprint arXiv:2411.04468},
  year         = {2024},
  url          = {https://doi.org/10.48550/arXiv.2411.04468}
}

@article{LiveLedger,
  author       = {Dayoon Ko and
                  Jihyuk Kim and
                  Sohyeon Kim and
                  Haeju Park and
                  Dahyun Lee and
                  Gunhee Kim and
                  Moontae Lee and
                  Kyungjae Lee},
  title        = {When Is Enough Not Enough? Illusory Completion in Search Agents},
  journal      = {arXiv preprint arXiv:2602.07549},
  year         = {2026},
  url          = {https://doi.org/10.48550/arXiv.2602.07549}
}

@article{ProactiveMemoryAgent,
  author       = {Yifan Wu and
                  Lizhu Zhang and
                  Yuhang Zhou and
                  Mingyi Wang and
                  Bo Peng and
                  Serena Li and
                  Xiangjun Fan and
                  Zhuokai Zhao},
  title        = {Remember When It Matters: Proactive Memory Agent for Long-Horizon
                  Agents},
  journal      = {arXiv preprint arXiv:2607.08716},
  year         = {2026},
  url          = {https://doi.org/10.48550/arXiv.2607.08716}
}

@article{LongHorizonHarness,
  author       = {Ziyu Ma and
                  Hailang Huang and
                  Shun Zou and
                  Yong Wang and
                  Shidong Yang and
                  Yiming Hu and
                  Fei Wei and
                  Xiangxiang Chu},
  title        = {LongHorizon-Harness: Advancing Long-Horizon Agents for Real-World
                  Tasks},
  journal      = {arXiv preprint arXiv:2608.01964},
  year         = {2026},
  url          = {https://doi.org/10.48550/arXiv.2608.01964}
}

@article{BM25,
  author       = {Stephen E. Robertson and
                  Hugo Zaragoza},
  title        = {The Probabilistic Relevance Framework: {BM25} and Beyond},
  journal      = {Found. Trends Inf. Retr.},
  volume       = {3},
  number       = {4},
  pages        = {333--389},
  year         = {2009},
  url          = {https://doi.org/10.1561/1500000019}
}

@inproceedings{SelfRefine,
  author       = {Aman Madaan and
                  Niket Tandon and
                  Prakhar Gupta and
                  Skyler Hallinan and
                  Luyu Gao and
                  Sarah Wiegreffe and
                  Uri Alon and
                  Nouha Dziri and
                  Shrimai Prabhumoye and
                  Yiming Yang and
                  Shashank Gupta and
                  Bodhisattwa Prasad Majumder and
                  Katherine Hermann and
                  Sean Welleck and
                  Amir Yazdanbakhsh and
                  Peter Clark},
  editor       = {Alice Oh and
                  Tristan Naumann and
                  Amir Globerson and
                  Kate Saenko and
                  Moritz Hardt and
                  Sergey Levine},
  title        = {Self-Refine: Iterative Refinement with Self-Feedback},
  booktitle    = {Advances in Neural Information Processing Systems 36: Annual Conference
                  on Neural Information Processing Systems 2023, NeurIPS 2023, New Orleans,
                  LA, USA, December 10 - 16, 2023},
  year         = {2023},
  url          = {http://papers.nips.cc/paper\_files/paper/2023/hash/91edff07232fb1b55a505a9e9f6c0ff3-Abstract-Conference.html}
}

@inproceedings{SelfNotes,
  author       = {Jack Lanchantin and
                  Shubham Toshniwal and
                  Jason Weston and
                  Arthur Szlam and
                  Sainbayar Sukhbaatar},
  editor       = {Alice Oh and
                  Tristan Naumann and
                  Amir Globerson and
                  Kate Saenko and
                  Moritz Hardt and
                  Sergey Levine},
  title        = {Learning to Reason and Memorize with Self-Notes},
  booktitle    = {Advances in Neural Information Processing Systems 36: Annual Conference
                  on Neural Information Processing Systems 2023, NeurIPS 2023, New Orleans,
                  LA, USA, December 10 - 16, 2023},
  year         = {2023},
  url          = {http://papers.nips.cc/paper\_files/paper/2023/hash/274d0146144643ee2459a602123c60ff-Abstract-Conference.html}
}

@misc{GPT54,
  title        = {Introducing {GPT-5.4} mini and nano},
  author       = {{OpenAI}},
  year         = {2026},
  month        = mar,
  howpublished = {OpenAI Official Blog},
  url          = {https://openai.com/index/introducing-gpt-5-4-mini-and-nano/},
}

@article{DeepSeekV4,
  author       = {DeepSeek{-}AI},
  title        = {DeepSeek-V4: Towards Highly Efficient Million-Token Context Intelligence},
  journal      = {arXiv preprint arXiv:2606.19348},
  year         = {2026},
  url          = {https://doi.org/10.48550/arXiv.2606.19348}
}

@misc{Gemini38,
  title        = {{Gemini 3.8 Flash} Model Card},
  author       = {{Google DeepMind}},
  year         = {2026},
  month        = sep,
  howpublished = {Google DeepMind Model Card},
  url          = {https://deepmind.google/models/model-cards/gemini-3-8-flash/},
}

\clearpage
\appendix

\definecolor{tabagent}{HTML}{F7E8CB}
\definecolor{tabenv}{HTML}{EBF3FA}
\newcommand{\cmark}{\textcolor{green!45!black}{\checkmark}}
\newcommand{\xmark}{\textcolor{red!70!black}{$\times$}}
\newcommand{\key}[1]{\textbf{\textcolor{black}{#1}}}
\newcommand{\turn}[1]{\textcolor{black!50}{#1:}}
\newcommand{\casebox}[2]{\par\vspace{1.5pt}\noindent{\setlength{\fboxsep}{2pt}\colorbox{#1}{\parbox[b]{\dimexpr\linewidth-2\fboxsep\relax}{\raggedright\ttfamily\fontsize{6.2}{7.2}\selectfont\color{black!60}#2\strut}}\vrule width 0pt depth \dimexpr\fboxsep+2.5pt\relax}\par\vspace{1.5pt}}
\makeatletter\setlength{\@fptop}{0pt}\makeatother
\newcommand{\taskbox}[1]{{\setlength{\fboxsep}{3pt}\colorbox{mygray}{\parbox[b]{\dimexpr\linewidth-2\fboxsep\relax}{#1\strut}}\vrule width 0pt depth \dimexpr\fboxsep+1.5pt\relax}}
\newcommand{\oursout}[1]{\multicolumn{1}{>{\columncolor{citeblue!10}[\tabcolsep][0pt]\raggedright\arraybackslash}p{\dimexpr0.57\linewidth-12pt\relax}@{}}{#1}}
\newcommand{\casesep}{\noalign{\vskip 0.25ex}\cdashline{1-3}\noalign{\vskip 0.5ex}}

\newtcolorbox{promptbox}[1]{colback=white, colframe=black!45, colbacktitle=mygray, coltitle=black, fonttitle=\footnotesize, title={#1}, boxrule=0.5pt, titlerule=0.5pt, arc=1.5mm, left=7pt, right=7pt, top=4pt, bottom=5pt, toptitle=2pt, bottomtitle=2pt, fontupper=\fontfamily{zi4}\selectfont\fontsize{7.2}{8.9}\selectfont\frenchspacing\raggedright}
\newcommand{\pline}[2][0pt]{\par\hangindent=\dimexpr#1+1.5em\relax\hangafter=1\noindent\hspace*{#1}#2}
\newcommand{\pskip}{\par\vspace{3pt}}
\newcommand{\psep}{\noalign{\vskip 4pt}}
\newcommand{\pslot}[1]{{\normalfont\itshape\{#1\}}}
\newcommand{\pgray}[1]{{\color{black!45}#1}}
\newcommand{\pnote}[1]{{\normalfont\itshape\color{black!55}#1}}
\newcommand{\pop}[1]{\textbf{#1}}
\newcommand{\pcode}[1]{{\fontfamily{zi4}\selectfont #1}}
\newcommand{\pparam}[1]{\par\hangindent=2.5em\hangafter=1\noindent\hspace*{1em}\pgray{$\bullet$}~#1}

\section{Qualitative Examples}
\label{sec:case_study}

In this section, we first show an environment-side tab of \modelname on one attempt (\cref{sec:tab_example}), and then compare DCI and \modelname with the same LLM on two other Workspace-Bench tasks (\cref{sec:case_full,sec:case_value}). Tool calls and observations are shown as recorded, shortened where marked.

\subsection{Example: An Environment-Side Tab of \modelname}
\label{sec:tab_example}

In contrast to \cref{fig:method}, which illustrates the environment-side tab on a hypothetical task, \cref{tab:tab_example} shows it on an actual Workspace-Bench task, as \modelname records it over one attempt. The task asks for a guide on business travel and reimbursement that covers five topics, from travel approval to violation handling, drawn from four policy files that the agent first finds by searching the workspace.

At the first turn, before opening any file, the agent decomposes the task into six requirements, one for each of the five topics that the task names and one for the guide itself, and adds them to the tab (A). The environment then fills the rest of the tab as the agent works (B): it records the 98 files that a search lists as candidates and the four policy files that the agent reads as the read-log entries L1 to L4, each holding the text the agent saw with its path and command. As each opened file moves from the candidates to the read log, the status line reports the change, from 98 candidates and no entries at turn 3 to 94 and four at turn 8.

The tab then connects each requirement to its source, as with the reimbursement standards (r3), whose source is L2, the entry of the policy file that opens with Article 7, on the standards for reimbursing travel expenses. When the agent lists its requirements at turn 10, the review shows r3 beside this line of L2, with an unopened file on the same subject (C), and after writing the guide at turn 12, the agent resolves r3 as \texttt{done} by citing L2 (D). At the stop at turn 22, the finish check confirms that all six are resolved before delivery (E).

Notably, the agent writes only two things to the tab over the whole task: the six requirements at the start and their resolutions before delivery. Everything else in \cref{tab:tab_example} (the read log, the candidates, the status line, the review, and the finish check) is kept and returned by the environment, which is what distinguishes the environment-side tab from a record that the model keeps for itself in its context.

\begin{table}[!htb]
    \centering
    \caption{The environment-side tab of \modelname over one attempt on a Workspace-Bench task, by the stages (A) to (E) of \cref{fig:method}. Boxes show what the agent writes to the tab (orange) and what the environment records or returns (blue), shortened with \ldots; key parts for requirement r3 are in bold.}
    \label{tab:tab_example}
    \vspace{-0.1in}
    \scriptsize
    \setlength{\tabcolsep}{3pt}
    \renewcommand{\arraystretch}{1.15}
    \renewcommand{\casesep}{\noalign{\vskip 0.25ex}\cdashline{1-2}\noalign{\vskip 0.5ex}}
    \begin{tabular}{@{} >{\raggedright\arraybackslash}p{0.13\linewidth} >{\raggedright\arraybackslash}p{\dimexpr0.87\linewidth-6pt\relax} @{}}
        \toprule
        \multicolumn{2}{@{}p{\linewidth}@{}}{\taskbox{\textbf{Task:} Integrate the core information from four business travel management policy documents, covering the full workflow of travel approval, travel requirements, reimbursement standards, reimbursement procedures, and violation handling. Ensure the guide is clear, easy to understand, and compliant with company policy, and generate \texttt{employee\_business\_travel\_and\_reimbursement\_guide.doc}.}} \\
        \midrule
        (A) Requirements & \turn{Turn 1} \texttt{add}, before any file is opened
        \casebox{tabagent}{r1: travel approval workflow \quad r2: travel requirements \quad \key{r3: reimbursement standards} \\ r4: reimbursement procedures \quad r5: violation handling \\ r6: generate employee\_business\_travel\_and\_reimbursement\_guide.doc} \\
        \casesep
        (B) Candidates & \turn{Turn 3} \texttt{find} lists 98 files, which the environment records as candidates; after turn 8, \texttt{list} returns
        \casebox{tabenv}{Unopened candidates: 94 (55 above threshold, marked *), ranked by relevance to the task; opening a file removes it \\ 1. \ldots/employee\_travel\_reimbursement\_policy\_2024.pdf * \\ 2. \ldots/travel\_expense\_reimbursement\_standards\_and\_attachment\_requirements.pdf * \\ \ldots} \\
        \casesep
        (B) Read log & \turn{Turns 5 to 8} \texttt{read} the four policy files, which the environment records as L1 to L4; \texttt{show} returns L2 as
        \casebox{tabenv}{L2: \ldots/business\_travel\_management\_policy\_2.txt (turn 6, 2286 chars) \\ command: read \ldots/business\_travel\_management\_policy\_2.txt \\ -{}-{}- \\ \key{Article 7 Standards for Reimbursement of Business Travel Expenses} \\ Employees traveling on business shall practice economy in transportation, meals, and \ldots \\ \ldots \\ Category City Category Accommodation Meal Allowance Long-distance Transportation \ldots \\ Manager level and above Tier-1 Cities 220 50 Bus or train (\ldots) 25 \\ \ldots} \\
        \casesep
        Status line & \turn{Turns 3 and 8} appended to each observation
        \casebox{tabenv}{{[}tab] \ldots{} | candidates: \key{98 unopened}, 59 above threshold | read log: \key{0 entries} | \ldots \\ {[}tab] \ldots{} | candidates: \key{94 unopened}, 55 above threshold | read log: \key{4 entries} | \ldots} \\
        \casesep
        (C) Review & \turn{Turn 10} \texttt{list}
        \casebox{tabenv}{Requirements: 6 open, 0 resolved \\ r1, r2 [open] \ldots \\ \key{r3 [open] reimbursement standards} \\ \hspace*{1em}from your reads: \\ \hspace*{2em}\key{L2 \ldots: Article 7 Standards for Reimbursement of Business Travel Expenses} \\ \hspace*{2em}\ldots \\ \hspace*{1em}never opened: \ldots/travel\_expense\_standards\_and\_reimbursement\_rules\_2024.docx * | \ldots \\ r4 to r6 [open] \ldots} \\
        \casesep
        (D) Resolution & \turn{Turns 15 to 20} \texttt{resolve}, after writing the guide at turn 12
        \casebox{tabagent}{r1 done (L1) \quad r2 done (L3) \quad \key{r3 done (L2)} \quad r4 done (L4) \quad r5 done (L4) \\ r6 unavailable (authored)} \\
        \casesep
        (E) Finish check & \turn{Turn 22} stops.
        \casebox{tabenv}{[tab] Finish check 1 of at most 1: you stopped. All 6 requirement items are resolved (5 done, 1 unavailable). Files you listed but never opened: 94 \ldots{} Check that each resolved item is actually in your deliverable \ldots}
        \turn{Turn 23} delivers the guide. \\
        \bottomrule
    \end{tabular}
    \vspace{-0.1in}
\end{table}

\subsection{Case Study 1: Tracking Every Source of a Deliverable}
\label{sec:case_full}

As the first case study, we compare DCI and \modelname with the same LLM on a task that asks for a water and electricity saving manual compiled from four policy files (\cref{tab:case_full}). Although both methods read the same four files within a few turns, they differ in what they keep track of. DCI writes the manual soon after reading the files, leaving the state of the task to its context window. In contrast, with \modelname, the agent first states one requirement per policy file (A), and the environment records each file it reads as a read-log entry (B); after writing the manual, the agent sees each requirement beside its matching entries in the review (C), resolves it against the entry of its own file (D), and receives a finish check confirming that none is left open (E). In the end, the manual of \modelname keeps all five chapters and twenty articles of the policy in their original order, whereas that of DCI merges them into an outline of its own and loses the chapter structure.

\begin{table}[!htb]
    \centering
    \caption{Case study on a Workspace-Bench task, comparing DCI and \modelname with the same LLM by the stages (A) to (E) of \cref{fig:method}. Boxes are colored as in \cref{tab:tab_example} and shortened with \ldots; the key parts are in bold.}
    \label{tab:case_full}
    \vspace{-0.1in}
    \scriptsize
    \setlength{\tabcolsep}{3pt}
    \renewcommand{\arraystretch}{1.15}
    \begin{tabular}{@{} >{\raggedright\arraybackslash}p{0.12\linewidth} >{\raggedright\arraybackslash}p{0.31\linewidth} >{\raggedright\arraybackslash}p{\dimexpr0.57\linewidth-12pt\relax} @{}}
        \toprule
        \multicolumn{3}{@{}p{\linewidth}@{}}{\taskbox{\textbf{Task:} Based on the four newly downloaded water and electricity policy files, prepare a quarterly water-saving and power-saving communication and inspection campaign, compile a complete \texttt{company\_water\_and\_electricity\_saving\_\allowbreak management\_manual.doc}, and use it for company-wide guidance and quarterly inspections. \ldots}} \\
        \midrule
        & \textbf{DCI} & \textbf{\modelname (Ours)} \\
        \midrule
        (A) Requirements & No record of the task. & \turn{Turn 6} \texttt{add}
        \casebox{tabagent}{\key{r1: Analyze water\_and\_electricity\_policy\_1.txt} \\ \key{r2 to r4: (the same for policy files 2 to 4)} \\ r5: Prepare a quarterly \ldots{} campaign plan \\ r6: Compile complete \ldots\_manual.doc in model\_output \\ r7: Output only the python list \ldots} \\
        \casesep
        (B) Reads & \turn{Turns 3 to 6} \texttt{read} the four files (Chapters 1 to 5, Articles 1 to 20). & \turn{Turns 7 to 10} \texttt{read} the same files, recorded as \key{L1 to L4}. \\
        \casesep
        Writing & \turn{Turn 10} writes the manual. & \turn{Turn 19} writes the manual. \\
        \casesep
        (C) Review & & \turn{Turn 23} \texttt{list}
        \casebox{tabenv}{r1 [open] Analyze water\_and\_electricity\_policy\_1.txt \\ \hspace*{1em}from your reads: \\ \hspace*{2em}\key{L1 \ldots policy\_1.txt: Company Water and Electricity Conservation Policy} \\ \hspace*{2em}\ldots \\ r2 to r7 [open] \ldots} \\
        \casesep
        (D) Resolution & & \turn{Turns 24 to 30} \texttt{resolve}
        \casebox{tabagent}{\key{r1 done (L1) \quad r2 done (L2) \quad r3 done (L3) \quad r4 done (L4)} \\ r5 to r7 unavailable (authored)} \\
        \casesep
        (E) Finish check & \turn{Turn 15} stops and delivers. & \turn{Turn 32} stops.
        \casebox{tabenv}{[tab] Finish check 1 of at most 1: you stopped. \key{All 7 requirement items are resolved} (4 done, 3 unavailable). Files you listed but never opened: 515 \ldots}
        \turn{Turns 33 to 36} opens the top listed file and delivers. \\
        \midrule
        Outcome & \xmark{} An outline of its own that merges Articles 1 and 2 into a purpose statement and Articles 19 and 20 into one bullet, losing the chapter structure, the article numbering, the publicity guide, and the inspection items. & \oursout{\cmark{} \key{The policy in its original order, Chapters 1 to 5 with Articles 1 to 20}, then the campaign plan, an inspection checklist, and rewards and penalties; each policy file is a requirement resolved against its own entry.} \\
        \bottomrule
    \end{tabular}
    \vspace{-0.1in}
\end{table}

\subsection{Case Study 2: Backing Every Row with Its Source}
\label{sec:case_value}

As the second case study, we consider a task that asks for a table summarizing ten purchase orders by supplier (\cref{tab:case_value}), where every order is expected in its own row. With DCI, the agent reads the orders one at a time, including that of Supplier 15 with the largest amount (\textyen 55,000), yet the table it writes holds a single row, so the suppliers it has read do not reach the deliverable. With \modelname, the environment records each order as a read-log entry as the agent reads it. Then, when the finish check reminds the agent that no requirement is recorded, the agent states one requirement per row, naming the supplier, status, amount, and source order, and resolves each against the entry of that order (such as the row of Supplier 15 against L12). In the end, the table of \modelname holds a row for every supplier, each tied in the tab to the order it comes from.

\begin{table}[!htb]
    \centering
    \caption{Case study on a second Workspace-Bench task, in the format of \cref{tab:case_full}: a value that DCI reads and leaves out of its table, and the same value in the tab, where the requirement for its row is resolved against the read-log entry of its purchase order. The gray box shows the purchase order as DCI reads it, and the blue box its read-log entry.}
    \label{tab:case_value}
    \vspace{-0.1in}
    \scriptsize
    \setlength{\tabcolsep}{3pt}
    \renewcommand{\arraystretch}{1.15}
    \begin{tabular}{@{} >{\raggedright\arraybackslash}p{0.12\linewidth} >{\raggedright\arraybackslash}p{0.31\linewidth} >{\raggedright\arraybackslash}p{\dimexpr0.57\linewidth-12pt\relax} @{}}
        \toprule
        \multicolumn{3}{@{}p{\linewidth}@{}}{\taskbox{\textbf{Task:} Based on the purchase order on the desktop, a table is generated by supplier, summarizing the amount and purchase type \ldots}} \\
        \midrule
        & \textbf{DCI} & \textbf{\modelname (Ours)} \\
        \midrule
        (B) Reads & \turn{Turn 13} \texttt{read} \ldots/Purchase\_Order\_8.txt
        \casebox{mygray}{Purchase Order \#1008 \\ \key{Supplier: Supplier 15} \\ \ldots \\ \key{Amount: \textyen 55000} \\ Status: Pending Approval} & \turn{Turn 7} \texttt{read} the same order; the environment records it as \key{L12}.
        \casebox{tabenv}{L12: \ldots/Purchase\_Order\_8.txt \\ \ldots{} \key{Supplier: Supplier 15} \ldots{} \key{Amount: \textyen 55000} \ldots} \\
        \casesep
        Writing & \turn{Turn 19} writes the table. & \turn{Turn 18} writes the table. \\
        \casesep
        (E) Finish check & \turn{Turn 21} stops and delivers. & \turn{Turn 21} stops.
        \casebox{tabenv}{[tab] Finish check 1 of at most 1: you stopped. \key{No requirement items were recorded. Add one per fact, figure, or deliverable element the task names} \ldots} \\
        \casesep
        (A) Requirements & & \turn{Turns 24 to 33} \texttt{add}
        \casebox{tabagent}{r2: Table row data: Supplier 2 | Stocked In | \textyen 16000 \\ \ldots{} (r3 to r8) \\ \key{r9: Table row data: Supplier 15 | Pending Approval | \textyen 55000. Source: \ldots/Purchase\_Order\_8.txt.} \\ \ldots} \\
        \casesep
        (D) Resolution & & \turn{Turn 39} \texttt{resolve}
        \casebox{tabagent}{r2 done (L8) \quad \ldots \quad \key{r9 done (L12)} \quad \ldots \quad r11 done (L9)} \\
        \midrule
        Outcome & \xmark{} A single row (\texttt{PH\_11 | 45000}): the orders read one by one, including that of Supplier 15, do not reach the table. & \oursout{\cmark{} \key{One row for each of the ten suppliers}, including \texttt{Supplier 15 | Pending Approval | 55000}, the largest amount; every row is tied in the tab to the entry of its order (r9 to L12).} \\
        \bottomrule
    \end{tabular}
    \vspace{-0.1in}
\end{table}

\clearpage
\section{Prompts}
\label{sec:prompts}

This section lists the prompts with which \modelname communicates with the agent, which are the same for every benchmark and every LLM. In each figure, values that the environment fills in at run time are shown in braces, and notes in gray italics are annotations that are not part of the prompts.

\begin{figure}[!hbp]
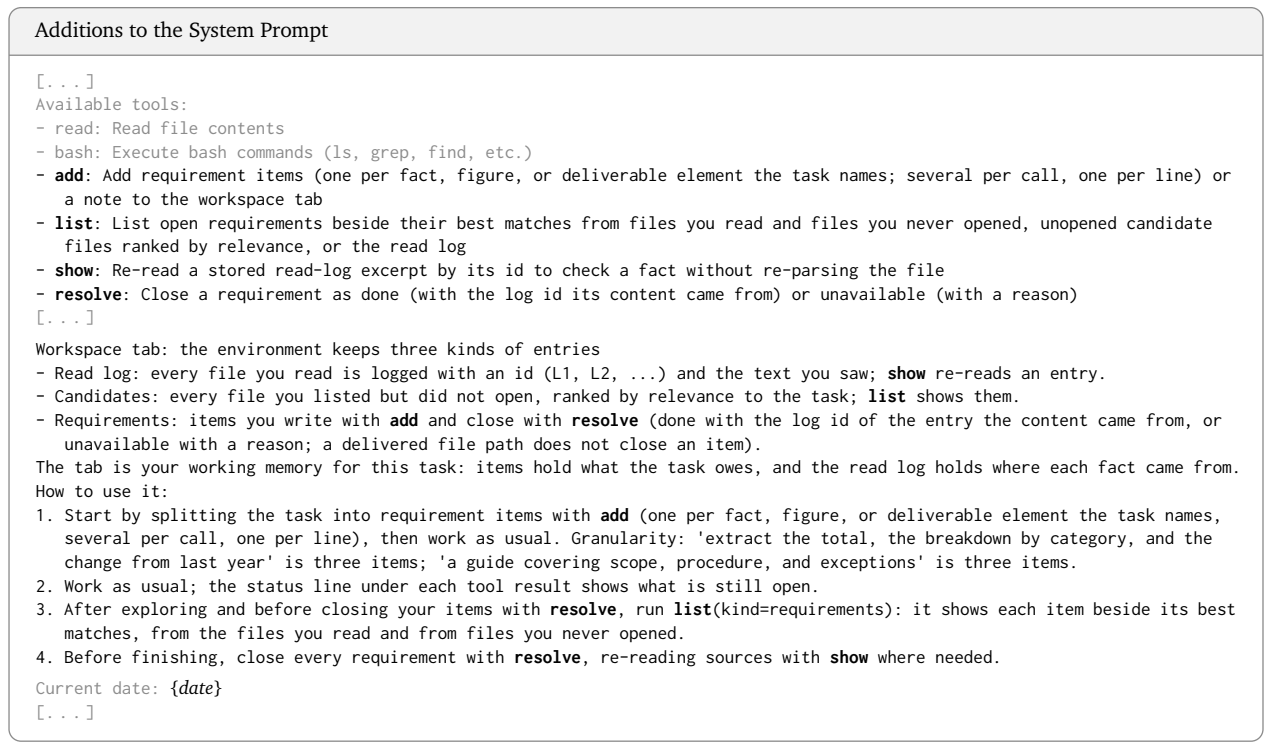

    \centering
    \begin{promptbox}{Additions to the System Prompt}
        \pline{\pgray{[\ldots]}}
        \pline{\pgray{Available tools:}}
        \pline{\pgray{- read: Read file contents}}
        \pline{\pgray{- bash: Execute bash commands (ls, grep, find, etc.)}}
        \pline{- \pop{add}: Add requirement items (one per fact, figure, or deliverable element the task names; several per call, one per line) or a note to the workspace tab}
        \pline{- \pop{list}: List open requirements beside their best matches from files you read and files you never opened, unopened candidate files ranked by relevance, or the read log}
        \pline{- \pop{show}: Re-read a stored read-log excerpt by its id to check a fact without re-parsing the file}
        \pline{- \pop{resolve}: Close a requirement as done (with the log id its content came from) or unavailable (with a reason)}
        \pline{\pgray{[\ldots]}}
        \pskip
        \pline{Workspace tab: the environment keeps three kinds of entries}
        \pline{- Read log: every file you read is logged with an id (L1, L2, ...) and the text you saw; \pop{show} re-reads an entry.}
        \pline{- Candidates: every file you listed but did not open, ranked by relevance to the task; \pop{list} shows them.}
        \pline{- Requirements: items you write with \pop{add} and close with \pop{resolve} (done with the log id of the entry the content came from, or unavailable with a reason; a delivered file path does not close an item).}
        \pline{The tab is your working memory for this task: items hold what the task owes, and the read log holds where each fact came from.}
        \pline{How to use it:}
        \pline{1. Start by splitting the task into requirement items with \pop{add} (one per fact, figure, or deliverable element the task names, several per call, one per line), then work as usual. Granularity: \textquotesingle extract the total, the breakdown by category, and the change from last year\textquotesingle{} is three items; \textquotesingle a guide covering scope, procedure, and exceptions\textquotesingle{} is three items.}
        \pline{2. Work as usual; the status line under each tool result shows what is still open.}
        \pline{3. After exploring and before closing your items with \pop{resolve}, run \pop{list}(kind=requirements): it shows each item beside its best matches, from the files you read and from files you never opened.}
        \pline{4. Before finishing, close every requirement with \pop{resolve}, re-reading sources with \pop{show} where needed.}
        \pskip
        \pline{\pgray{Current date: }\pslot{date}}
        \pline{\pgray{[\ldots]}}
    \end{promptbox}
    \vspace{-0.1in}
    \caption{Additions of \modelname to the system prompt (\cref{sec:keeping}). Text in gray is the system prompt of the harness, shown only to locate the additions, with [\ldots] marking omitted parts.}
    \label{fig:prompt_system}
\end{figure}

\paragraph{System Prompt and Operations}
\cref{fig:prompt_system} shows what \modelname adds to the system prompt: a one-line description of each operation and a block that asks the agent to state the requirements at the start, review them after exploring, and resolve them before finishing. \cref{fig:prompt_operations} then shows the four operations on the tab, as given to the LLM as tools with their descriptions and parameters.

\paragraph{Messages during the Task}
\cref{fig:prompt_messages} shows the messages that the environment returns while the agent works: the status line, the review of the requirements, and the replies of \texttt{resolve}.

\paragraph{Finish Check}
\cref{fig:prompt_finish} shows the finish check, which the environment sends once at the first stop, with content that depends on whether the requirements are missing, open, or all resolved.

\begin{figure}[tbp]
    \centering
    \begin{promptbox}{Operations on the Tab}
        \begin{tabular}{@{}>{\raggedright\arraybackslash}p{0.09\linewidth}@{\hspace{6pt}}>{\raggedright\arraybackslash}p{\dimexpr0.91\linewidth-6pt\relax}@{}}
            \pop{add} &
            Add items to the workspace tab. kind=requirement: one item per fact, figure, or deliverable element the task names (each number to extract, each table, section, or file to produce), not one per sentence; write them at the start; several items can go in one call, one per line; returns their ids (r1, r2, ...). kind=note: free text to remember (returns n1).
            \pparam{kind: requirement | note}
            \pparam{text: the item text} \\
            \psep
            \pop{list} &
            List tab contents. kind=requirements: each open requirement item beside its best matches, the matching line of each file you read and the files you never opened (resolved ones with the source line they were closed against). kind=candidates: files you listed but never opened, ranked by relevance to the task, most relevant first. kind=log: read-log entries with ids and anchors, newest first. Long lists are paged (page=2, ...).
            \pparam{kind: requirements | candidates | log}
            \pparam{page \pnote{(optional)}: page number, default 1} \\
            \psep
            \pop{show} &
            Re-read one read-log entry by id (L3): returns its anchor (file path and the command that produced it) and the stored text as you first saw it, so a fact can be checked without parsing the file again.
            \pparam{id: a read-log id such as L3} \\
            \psep
            \pop{resolve} &
            Close a requirement item. status=done with ref=<read-log id>, the entry its content came from; status=unavailable with ref=<reason>: computed (derived from sources, not stored), authored (you had to write it), or not in the workspace.
            \pparam{id: requirement id such as r3}
            \pparam{status: done | unavailable}
            \pparam{ref: done: the read-log id the content came from (L2); unavailable: the reason} \\
        \end{tabular}
    \end{promptbox}
    \vspace{-0.1in}
    \caption{Operations on the tab (\cref{sec:tab}), each given to the LLM as a tool with a description, followed by its parameters (marked $\bullet$) with the values allowed or a description of each parameter.}
    \label{fig:prompt_operations}
\end{figure}

\begin{figure}[tbp]
    \centering
    \begin{promptbox}{(a) Status Line}
        \pline{[tab] requirements: \pslot{k} open (\pslot{open ids}), \pslot{m} resolved | candidates: \pslot{u} unopened, \pslot{s} above threshold | read log: \pslot{n} entries | last tab access: \pslot{t} turns ago | ops: \pop{add} \pop{list} \pop{show} \pop{resolve}}
        \pline{\pnote{Before the first requirement is stated, the first field reads:} requirements: none added yet}
    \end{promptbox}
    \vspace{3pt}
    \begin{promptbox}{(b) Review}
        \pline{Requirements: \pslot{k} open, \pslot{m} resolved}
        \pline{\pslot{id} [open] \pslot{requirement}}
        \pline[1.5em]{from your reads:}
        \pline[3em]{\pslot{log id} \pslot{path}: \pslot{matching line}\hfill\pnote{up to five entries}}
        \pline[1.5em]{never opened: \pslot{path} | \pslot{path} | ...\hfill\pnote{up to five files}}
        \pline{\pslot{id} [done \pslot{log id}] \pslot{requirement}}
        \pline[1.5em]{closed against \pslot{log id} \pslot{path}: \pslot{matching line}}
        \pline{\pslot{id} [unavailable \pslot{reason}] \pslot{requirement}}
        \pline{\pslot{u} files never opened in total: \pop{list}(kind=candidates)}
        \pline{\pnote{Before the first requirement is stated:} No requirement items yet. Add one per thing the task asks for with \pop{add}(kind=requirement).}
    \end{promptbox}
    \vspace{3pt}
    \begin{promptbox}{(c) Replies of \pcode{resolve}}
        \pline{\pnote{Accepted:} \pslot{id} \pslot{status} (\pslot{ref}). Open: \pslot{k}.}
        \pline{\pnote{Unknown requirement:} No requirement \pslot{id}. Open ids: \pslot{open ids}.}
        \pline{\pnote{Citation that is neither an entry nor a file:} \pslot{id} stays open: ref must be an existing read-log id (known: \pslot{log ids}).}
        \pline{\pnote{Citation of a file in the output folder:} \pslot{id} stays open: a model\_output path does not close an item in this configuration. Point at the read-log entry the content came from (ref=L<id>; \pop{list}(kind=log) lists them), or close it as unavailable with the reason (computed, authored, or not in the workspace).}
        \pline{\pnote{Citation of a file not in the read log:} \pslot{id} stays open: \pslot{ref} is not in the read log. Open the file first, then close with its log id, or close the item as unavailable with the reason.}
        \pline{\pnote{Citation that shares no term with the requirement:} \pslot{id} stays open: the stored excerpt of \pslot{ref} does not mention the item content. Point at the entry the content actually came from (\pop{list}(kind=log), \pop{show}(id)), or close the item as unavailable with the reason.}
        \pline{\pnote{Unavailable without a reason:} \pslot{id} stays open: give the reason as ref (computed, authored, or not in the workspace).}
    \end{promptbox}
    \vspace{-0.1in}
    \caption{Messages from the environment during the task: (a) the status line, appended to each observation (\cref{sec:keeping}); (b) the review, returned by \texttt{list} with kind=requirements; and (c) the replies of \texttt{resolve} (\cref{sec:settling}). In the review, entries and files are ranked by BM25 against the requirement, and each entry shows its first line sharing a term with it.}
    \label{fig:prompt_messages}
\end{figure}

\begin{figure}[tbp]
    \centering
    \begin{promptbox}{Finish Check}
        \pline{[tab] Finish check 1 of at most 1: you stopped. \pslot{state of the requirements}}
        \pskip
        \pline{\pnote{(a) If no requirement was stated:}}
        \pline[1.5em]{No requirement items were recorded.}
        \pline[1.5em]{Add one per fact, figure, or deliverable element the task names with \pop{add}(kind=requirement), check each against the sources and your deliverable, then resolve them.}
        \pline{\pnote{(b) If some requirements are open:}}
        \pline[1.5em]{\pslot{k} requirement items are still open:}
        \pline[3em]{\pslot{id} \pslot{requirement}\hfill\pnote{one line per open requirement}}
        \pline[1.5em]{Resolved: \pslot{d} done, \pslot{v} unavailable.}
        \pline{\pnote{(c) If all requirements are resolved:}}
        \pline[1.5em]{All \pslot{n} requirement items are resolved (\pslot{d} done, \pslot{v} unavailable).}
        \pskip
        \pline{Files you listed but never opened: \pslot{u} (\pslot{s} above the relevance threshold, marked *); the 10 most relevant to the task:}
        \pline[1.5em]{\pslot{path} *\hfill\pnote{one line per file, ranked by BM25 against the task}}
        \pline[1.5em]{... \pslot{u $-$ 10} more: \pop{list}(kind=candidates)}
        \pline{Read log: \pslot{n} entries (L1 to L\pslot{n}); \pop{show}(id) re-reads one, \pop{list}(kind=log) lists them.}
        \pskip
        \pline{\pnote{In (a) and (b):}}
        \pline[1.5em]{For each open requirement: add the missing content to your deliverable (re-read the source with \pop{show} or the file itself) and close it with \pop{resolve}(id, status=done, ref=<log id>), or close it with \pop{resolve}(id, status=unavailable, ref=<reason: computed, authored, or not in the workspace>). Then finish again as the task instructs.}
        \pline{\pnote{In (c):}}
        \pline[1.5em]{Check that each resolved item is actually in your deliverable (\pop{show} re-reads a source excerpt), open a listed file if it looks like a source you missed, then finish again as the task instructs.}
    \end{promptbox}
    \vspace{-0.1in}
    \caption{Finish check, sent once at the first stop (\cref{sec:settling}). A listed file is above the relevance threshold, and marked *, when its path shares at least two distinct terms with the text of the task.}
    \label{fig:prompt_finish}
\end{figure}

\end{document}